\pdfoutput=1

\documentclass[11pt]{article}

\usepackage[final]{acl}

\usepackage{times}
\usepackage{latexsym}

\usepackage[T1]{fontenc}

\usepackage[utf8]{inputenc}

\usepackage{microtype}

\usepackage{inconsolata}

\usepackage{booktabs}
\usepackage{pgfplots}
\usepackage{pgf-pie}
\usepackage{graphicx}
\usepackage{svg}
\usepackage{threeparttable}
\usepackage{caption}
\usepackage{subcaption}
\usepackage{hyperref}
\usepackage{multirow}

\usepackage{siunitx}
\pgfplotsset{compat=1.18}
\usepgfplotslibrary{colorbrewer}
\usepgfplotslibrary{statistics}
\graphicspath{{assets/}}

\title{Investigating Wit, Creativity, and Detectability of Large Language Models in Domain-Specific Writing Style Adaptation of Reddit's Showerthoughts}

\author{Tolga Buz\thanks{Equal contribution}, Benjamin Frost\footnotemark[1], Nikola Genchev\footnotemark[1], \\ \textbf{Moritz Schneider}, \textbf{Lucie-Aimée Kaffee}, \textbf{Gerard de Melo} \\
  Hasso Plattner Institute / University of Potsdam, Germany \\
  \texttt{tolga.buz@hpi.de, gdm@demelo.org}}

\begin{document}
\maketitle
\begin{abstract}
Recent Large Language Models (LLMs) have shown the ability to generate content that is difficult or impossible to distinguish from human writing. 
We investigate the ability of differently-sized LLMs to replicate human writing style in short, creative texts in the domain of \emph{Showerthoughts}, thoughts that may occur during mundane activities.
We compare GPT-2 and GPT-Neo fine-tuned on Reddit data as well as GPT-3.5 invoked in a zero-shot manner, against human-authored texts. 
We measure human preference on the texts across the specific dimensions that account for the quality of creative, witty texts.
Additionally, we compare the ability of humans versus fine-tuned RoBERTa classifiers to detect AI-generated texts.
We conclude that human evaluators rate the generated texts slightly worse on average regarding their creative quality, but they are unable to reliably distinguish between human-written and AI-generated texts. We further provide a dataset for creative, witty text generation based on Reddit Showerthoughts posts.
\end{abstract}

\section{Introduction}

As Large Language Models (LLMs) continue to advance, it becomes increasingly challenging for humans to distinguish AI-generated and human-written text.
Generated text may appear surprisingly convincing, %
inciting debates whether new forms of evaluating models are necessary~\citep{sejnowski2023large}.
The high quality of LLM outputs can benefit diverse use cases, while also increasing the risk of enabling more sophisticated spam, misinformation, and hate speech bots \cite{ManduchiEtAl2024GenAI}. LLMs are known to master various aspects of grammar and basic semantics.
Yet, one goal that still has proven non-trivial using LLMs
is that of generating creative text \cite{DBLP:journals/corr/abs-2309-14556}, especially in the realm of humour \citep{jentzsch2023chatgpt}. 

We seek to understand the ability of differently-sized LLMs to replicate human writing style in short and creative texts as shared in the \textit{Showerthoughts} community on Reddit, which exhibits humour, cleverness, and creativity -- often in a single sentence.
The Showerthoughts community (Reddit's 11th largest) provides a unique dataset of short texts with a characteristic writing style drawing from general creative qualities.
To understand how well models of different sizes can replicate such witty Reddit posts, we fine-tuned two LLMs, GPT-2 (Medium) and GPT-Neo, on posts from this online community. 
Additionally, we used GPT-3.5-turbo as a zero-shot model, i.e., without additional fine-tuning for our specific task.
We evaluated how well the AI-generated texts emulate the style of Showerthoughts. 
To this end, we employed a mixed-method approach:
We compare genuine, human-authored posts with generated Showerthoughts based on various lexical characteristics as well as in their similarity in sentence embeddings. 
Furthermore, we conducted a human evaluation study to assess the human evaluators' perception of the creative quality (specifically,  logical validity, creativity, humour, and cleverness) and to measure how easily AI-generated texts can be detected. 

We find that participants cannot reliably detect AI-generated texts, as the LLMs come close to human-level quality.
Generating humour remains a challenging task, but shows a promising future for the generation of short, witty, and creative statements. 
We find that a machine learning (ML) classifier, trained on Showerthoughts, succeeds at robustly distinguishing human-authored from AI-written text.
Thus, there remains potential for current AI-generated content to be identified, even in the ambiguous realm of humour and creative text.

We summarize our contributions in this paper as follows: 
(1) A new dataset for creative, witty text generation based on Reddit Showerthoughts posts.\footnote{Dataset accessible via \href{https://github.com/aiintelligentsystems/showerthoughts-dataset}{our GitHub repository.}}
(2) Experiments with three different models for the generation of creative, witty text. 
(3) Evaluation of human perception of creative language generation through a survey. 
(4) Experiments on automated authorship identification of the text as human-written or AI-generated.

\section{Background and Related Work}

\paragraph{Reddit and Showerthoughts}
Reddit is a social media platform that %
is organized in communities called \emph{subreddits}, which
exist for a plethora of topics %
-- all written, curated, voted, and commented on by the community. 
This provides a diverse and valuable research subject; each subreddit is characterized by a distinct writing style and type of content \cite{AgrawalBuzDeMelo2022WallStreetBetsSocial,BuzSchneiderKaffeeDeMelo2024WSBPredictions}. %

Our work is centered on the r/Showerthoughts subreddit\footnote{\href{https://www.reddit.com/r/Showerthoughts}{\texttt{www.reddit.com/r/Showerthoughts}}}, which defines \textit{Showerthought} as ``a loose term that applies to the types of thoughts you might have while carrying out a routine task like showering, driving, or daydreaming. At their best, Showerthoughts are universally relatable and find the amusing/interesting within the mundane.''
In general, popular Showerthoughts exhibit %
wit (or cleverness), creativity, and sometimes humour, which come from the realization of matters that lie in everyday life's banality, which are well thought out but tend to go unnoticed.
They condense various intellectual qualities into short texts that often allude to a deeper context -- these qualities can be facilitators of a text's success in various other settings, including posting on social media or copy-writing for marketing purposes.
One of the community's most successful post goes as follows: ``When you're a kid, you don't realize you're also watching your mom and dad grow up.''\footnote{Accessible via \url{https://www.reddit.com/awd10u/}}

To the best of our knowledge, there is only one related paper focused on Showerthoughts, which covers a neuro-scientific perspective \cite{crawford2020daydreaming}. 
Limited research exists that uses Showerthoughts data among other subreddits, but on completely different topics, e.g., detection of suicidal thoughts \cite{aladaug2018detecting}, predicting conversations \cite{kim2023predicting}, or changes of the community \cite{lin2017better}. 
Our work is the first to analyse the texts that are shared in this community from a perspective of computational linguistics and the first to publish a Showerthoughts dataset.

\paragraph{Creative Quality in Natural Language Generation}

Early work on computational creativity found that while computers can aid in the creative process, it has long remained difficult to achieve novelty and quality with such systems \citep{gervas2009computational}.
More recent LLMs possess a remarkable ability to produce entirely novel content, but \citet{chakrabarty2022s} find that they have limited capabilities w.r.t.\ figurative language, and that full stories generated by LLMs seem to be of far inferior quality compared to those written by professional authors \cite{DBLP:journals/corr/abs-2309-14556}.
Further, popular LLMs such as ChatGPT have been found to be subpar at writing creative and humourous content such as jokes \cite{jentzsch2023chatgpt}.
For many creative tasks, such as writing convincing poems, human intervention may be needed to create high-quality text \cite{chakrabarty2022help}, and the temperature hyperparameter may have a significant impact on the creativity of LLM-generated texts \cite{davis2024temperature}.
AI-assisted writing may lead to improved results \cite{roemmele2021inspiration} and LLMs have been perceived as writing collaborators by professional writers \cite{chakrabarty2023creativity}.
However, it is yet to be seen how the generation of creative, witty text without human intervention can be improved to agree with human preferences.

\paragraph{Authorship Identification}
There have been significant advancements in LLMs generating grammatically correct sentences adhering to semantic rules, even purportedly attaining human levels~\citep{kobis2021-artificial, clark2021-all}.
This presents opportunities in areas such as accessibility of information and education, and enhanced productivity~\citep{dwivedi2023so,noy2023experimental}. 
However, it also poses a threat to the credibility of %
information~\citep{kreps2022-all,kumar2018false}, especially as social media users often fail to detect bots~\citep{kenny2022duped}, while such bots continue to evolve and spread~ misinformation~\citep{abokhodair2015dissecting, shao2018spread}.
Indeed, \citet{ippolito2020-automatic} found that even trained participants struggle to identify AI-generated texts. 
\citet{kobis2021-artificial} further found that while completely random texts could be detected, cherry-picked texts could not be distinguished by humans.
The model size used to generate texts affects participants' performance -- both studies used %
smaller models 
(GPT-2 with 355M, 774M, and 1.5B parameters, respectively), whereas participants confronted with never models such as GPT-3 performed significantly worse in a similar study \citep{clark2021-all, brown2020-language}.
With larger model sizes, humans require more time to decide, and their accuracy declines~\cite{brown2020-language}. 
In very recent work, \citet{chen2024llmgenerated} find that LLM-generated misinformation can be more deceptive than when written by human authors, and \citet{munozortiz2023contrasting} identify measurable differences between AI-generated and human-written texts.

As LLMs advance rapidly, it becomes crucial to understand what type of generated content humans can detect and how to detect generated content automatically.
For automatic authorship identification, \citet{wani2017sneak} use ML classifiers to detect bots.
\citet{ippolito2020-automatic} use a fine-tuned BERT-based binary classifier to label texts as human-written or AI-generated. 
However, their model lacks generalizability -- when trained on top-$k$ samples and evaluated on non-truncated random samples, the model only achieves 43.8\% accuracy.
The sharp increase in discussions about misuse and plagiarism using tools such as ChatGPT has shifted researchers' focus on this area, e.g., \citet{mitchell2023-detectgpt} proposed DetectGPT, a zero-shot model for detecting AI-generated text, and \citet{deng2023efficient} proposed a Bayesian Surrogate Model, claiming to outperform DetectGPT. 
\citet{tang2024science} provide an overview of further detection techniques.

\section{Data Compilation}
\label{subsec:dataset}

To create the Showerthoughts dataset, we used the publicly available Pushshift API~\citep{clark2021-all, brown2020-language} to extract submissions from the Showerthoughts subreddit from April 2020 to November 2022, resulting in an initial collection of 1.3 million posts.\footnote{In mid 2023, Reddit changed their API guidelines, forcing Pushshift to restrict its access to Reddit moderators only. Our datasets were collected before this change occurred.}
We discard posts that have been deleted or removed (often due to rule violation) as well as those that contain images or additional explanations in their body text (as the community's rules require the full Showerthought to be contained in the title).
Accordingly, we only use each post's title for our experiments, resulting in a dataset of 411,189 Showerthoughts.
An analysis of the most frequent choices of words reveals that they are often about people, life, common objects, and the world in general. A frequent word analysis indicates that they often compare things using, e.g., ``more'', ``other'', ``old'', ``good''.

In order to obtain a ground truth about the lexical characteristics of the dataset and later compare them with the generated texts, we conducted several tests on 5,000 randomly selected examples, focusing on sentence complexity, length, grammar, and vocabulary, the results of which are summarized in Table~\ref{table:lexical_comparison} (in the first row `Genuine').
The complexity score is based on the Flesch-Kincaid grade level, which quantifies a text's complexity based on the number of words per sentence and syllables per word~\citep{kincaid1975derivation}. 
For example, a score of 7.0 indicates that a 7\textsuperscript{th}-grade student (or a person with at least seven years of education) would typically be able to read and understand the respective text.\footnote{Tests are conducted with the {\tt textstat}, {\tt language\_tool\_python}, and {\tt nltk} libraries.}

\section{Experimental Setup}

\begin{figure}[htbp]
  \centering
  \includegraphics[width=0.8\linewidth]{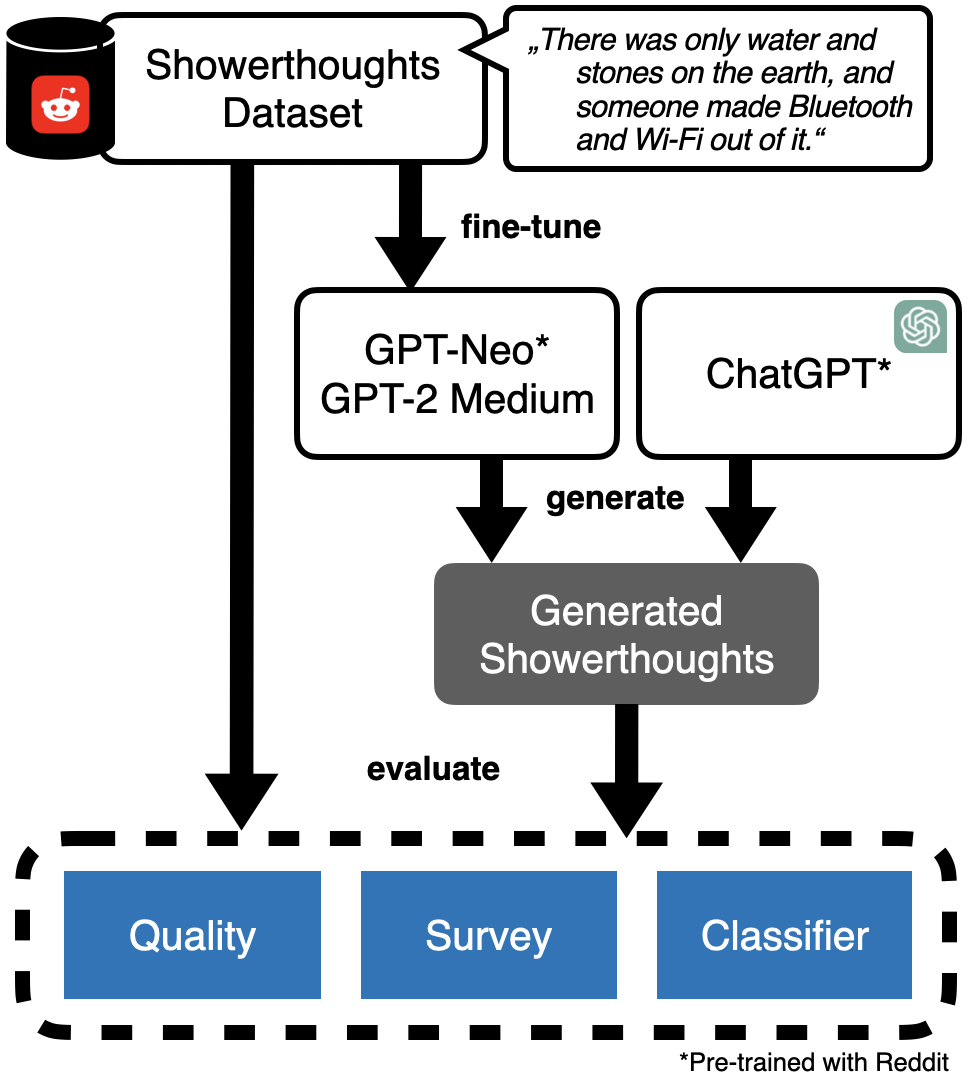}
  \caption{\centering Overview of our experimental setup}
  \label{figure:overview}
\end{figure}

In the following, we detail our experimental setup for addressing our three research questions. 
We explain our process for generating %
Reddit Showerthoughts-like texts with differently sized selected LLMs.
These texts are subsequently evaluated through a survey, assessing several textual aspects.
Additionally, we compare the ability of humans and fine-tuned BERT-based classifiers in detecting originality.
An overview of this experimental setup is given in Figure~\ref{figure:overview}.

\subsection{LLM Fine-Tuning and Prompting}
\label{subsec:model_finetuning}
We consider two setups for the generation of Showerthoughts; (1) two models of different sizes are fine-tuned; (2) ChatGPT (based on GPT-3.5-turbo) is invoked to generate Showerthoughts in a zero-shot setting.

\paragraph{Fine-tuning GPT-2 and GPT-Neo}
\label{subsec:gpt2}
\label{subsec:gpt_neo}
For the fine-tuned models, we select GPT-2 Medium (355M parameters) and GPT-Neo (2.7B parameters) and fine-tune them on the aforementioned Showerthoughts dataset.
To later be able to induce the models to generate Showerthoughts, each instance is wrapped around two previously unseen tokens, {\tt <|showerthought|>} and {\tt <|endoftext|>}. These serve as prompt and end-of-text markers, respectively, during generation.
We use the standard parameters for text generation for both models, including a temperature value of 0.9.

GPT-2 Medium\footnote{Accessible via \url{https://huggingface.co/gpt2-medium}} is a unidirectional causal language model that generates text sequences, using 355 million parameters.
This was the smallest LLM still able to generate sensible results in our initial evaluation during LLM selection.
We use AdamW for optimization, the {\tt GPT2Tokenizer}, a maximum learning rate of $3 \times 10^{-5}$ with 5,000 warm-up steps, a batch size of 16, and train the model for five epochs on the task of next token prediction. 

GPT-Neo is an architecturally upgraded model compared to GPT-2 that closely resembles GPT-3, with 2.7 billion parameters and trained on the Pile dataset~\citep{gao2020pile}. 
We selected the same hyperparameters as for GPT-2 besides using Adafactor optimization, which provides manual control over the learning rate and has better memory efficiency~\citep{shazeer2018adafactor}. We used a learning rate of $2 \times 10^{-5}$, which is reduced to $7 \times 10^{-6}$ over five epochs, and a batch size of 32.

\paragraph{Zero-shot Text Generation with ChatGPT}

In initial experiments, we found that a basic prompt (\textit{``Please generate 10 Showerthoughts"}) results in repetition of content and structure in generated texts, in accordance with the findings of \citet{jentzsch2023chatgpt}.
We therefore extended the prompt by including a definition of Showerthoughts, alongside instructions for enhancing wit, creativity, and humour, and varying sentence structure. 
This resulted in the following prompt:

    \textit{"Please generate 100 Showerthoughts, which are inspired by the Reddit community r/Showerthoughts. Vary the sentence structure between the different sentences, and try to be clever, creative, and funny. The Showerthoughts should be relatable and connected to things that people might encounter during mundane tasks."}

This process was repeated 50 times to sample a total of 5,000 Showerthoughts. We use the standard settings for text generation, including a temperature value of 0.7.

\subsection{Survey of Human Preferences}
\label{subsec:survey}

We evaluated the results of the text generation models by means of a survey.
The participants were randomly split into two groups to evaluate a larger number of Showerthoughts while ensuring an adequate number of responses per Showerthought and a reasonable completion time (around 25 minutes).
Each group evaluated 15 human-written and ten AI-generated Showerthoughts, each from GPT-2, GPT-Neo, and ChatGPT.
Participants were not informed about the distribution of the sources and received the texts in a random order to prevent evaluation bias.
The Showerthoughts were selected randomly and manually filtered to exclude posts harboring vulgarity or a ``not safe for work'' (NSFW) topic.

The survey starts with a briefing on Reddit and r/Showerthoughts, and we informed participants that they will evaluate 45 Showerthoughts, some of which are written by humans and some generated by LLMs.
We further ask demographic questions, %
including age group and the level of experience with Reddit, Showerthoughts, and Machine Learning on a five-point scale. 
Then, participants were asked to evaluate a series of 45 Showerthoughts by rating along six dimensions (each on a six-point Likert scale): (1) ``I like this Showerthought'', (2) ``It makes a true/valid/logical statement'', (3) ``It is creative'', (4) ``It is funny'', (5) ``It is clever'', and (6) ``I believe this Showerthought has been written by a real person''. 
These criteria were selected to capture the quality of a Showerthought from diverse angles, and are also applicable to comparable short texts such as social media posts and marketing texts.
For evaluation, we consider the average scores of the selected Likert scale from 1 (lowest) to 6 (highest). 
This method is widely used, e.g., in \citet{tang2021investigating}.
Finally, the participants could optionally provide a free-text explanation or reasoning on how they decided. %

\subsection{Authorship Identification}
\label{subsec:detection_with_roberta}

As a counterpart to the human evaluators on the task of authorship identification, we fine-tuned a total of four RoBERTa-based models\footnote{Specifically: \textit{RoBERTaForSequenceClassification}.} \cite{liu2019roberta} for binary classification of each input Showerthought as either human-written or AI-generated.
For the training and testing of the three LLM-specific RoBERTa classifiers, we used 10,000 randomly selected Showerthoughts per class (i.e., genuine, generated) for GPT-2 and GPT-Neo, and 5,000 examples for the ChatGPT version (due to the smaller generated dataset size). 
In addition, we trained and tested another RoBERTa classifier on a combined set of 15,000 examples per class (i.e., 5,000 per LLM source).
All datasets were randomly split at a 80--20 ratio for training and testing.
We assessed the classifiers in three setups; (1) evaluating 
the three LLMs' outputs compared to human-written (genuine) text separately; (2) evaluating all three LLMs' outputs combined compared to human-written text; (3) training the classifier on one LLM's outputs (GPT-Neo) and evaluating it on another LLM's outputs (GPT-2, ChatGPT, and all combined).

All versions of the classifier were trained with the tokenizer of RoBERTa-Base, AdamW optimization, a learning rate of $2 \times 10^{-5}$, batch size of 32, and a linear scheduler with 300 warm-up steps. 
To compute the loss for a given prediction, the model receives the tokenized Showerthought and the corresponding label indicating whether the Showerthought was genuine or generated.

\section{Results}

This section presents our experimental results.
Section~\ref{subsec:statistical_characteristics} compares %
lexical characteristics, showing that the LLMs come %
close to human quality. %
Next, Section~\ref{subsec:survey_results} explores the survey results, %
providing insights into crucial Showerthought attributes such as logical validity and creativity.
Lastly, Section~\ref{subsec:roberta_results} reports %
on our authorship identification, including patterns to distinguish between human-written and AI-generated Showerthoughts.

\subsection{Characteristics of Generated Showerthoughts}
\label{subsec:statistical_characteristics}

To assess the quality and similarity of generated to original Showerthoughts,
we apply the linguistic metrics described in Section~\ref{subsec:dataset} to the AI-generated Showerthoughts utilizing 5,000 random samples per source (for ChatGPT we use all 5,000 texts generated). %
Table~\ref{table:lexical_comparison} shows that human-written (genuine) Showerthoughts have a larger vocabulary, 
are slightly more complex, and contain more difficult words and grammar mistakes. 
Based on these metrics, GPT-Neo's generated texts 
are closer to genuine texts
compared to the significantly smaller GPT-2. %
ChatGPT 
ranks closest to the human reference regarding average complexity and length, slightly behind GPT-Neo regarding vocabulary size, but farthest away from the reference in terms of difficult words and grammar mistakes. 
We find that the models %
produce a negligible amount of duplicate Showerthoughts (GPT-2: 13 of 10,000, GPT-Neo: 162 of 10,000, ChatGPT: 6 of 5,000). %

\begin{table}[htbp]
    \centering
    \scalebox{0.85}{
    \begin{threeparttable}
    \begin{tabular}{l|cccc}
        \toprule
        Source &                Genuine              &   GPT-Neo     &   GPT-2       &   ChatGPT     \\
        Compl.\tnote{1} &   7.4 \small$\pm$ 3.4 & 6.8 \small$\pm$ 3.0   & 6.3 \small$\pm$ 2.7 & 6.9 \small$\pm$ 2.4 \\
        Length\tnote{1} &       81 \small$\pm$ 38   & 88 \small$\pm$ 39     & 87 \small$\pm$ 33   & 81  \small$\pm$ 21 \\
        Vocab.\tnote{2} &   13,000             &  8,700               & 4,900              & 7,200              \\
        Diffic.\tnote{3} &   1.2                 & 0.7                   & 0.4                 & 0.36                \\
        Errors\tnote{3} &     0.3                 & 0.2                   & 0.1                 & 0.05                \\
        \bottomrule
    \end{tabular}
    \begin{tablenotes}
        \item[1] Mean linguistic complexity (Flesch-Kincaid grade level) and length with standard deviation.
        \item[2] Vocabulary size in number of unique words.
        \item[3] Number of difficult words and grammatical errors per sentence.
    \end{tablenotes}
    \end{threeparttable}
    }
    \caption{Comparison of common lexical characteristics (based on 5,000 random samples per source)}
    \label{table:lexical_comparison}
\end{table}

\paragraph{Comparison of Sentence Embeddings}

How semantically diverse are Showerthoughts and are our LLMs able to match this diversity?
To answer this, we employ sentence embeddings\footnote{SBERT embeddings in their default, pre-trained configuration ({\tt all-MiniLM-L6-v2})}  %
for comparing the similarity between human-written and AI-generated %
content, and to measure the linguistic distance to texts from other subreddits.
We have reviewed the embeddings of 1,000 randomly sampled Showerthoughts per source visualized with the t-SNE algorithm~\citep{van2008visualizing}; %
GPT-2 and GPT-Neo produce more diverse texts than zero-shot ChatGPT, which matches human-written Showerthoughts based on their output distributing across the same semantic clusters as the human-written texts (Figure~\ref{figure:embeddings_comparison}, in Appendix).
When comparing these embeddings to 1,000 randomly selected titles from different, similarly large and popular subreddits, we find that every subreddit has a distinct focus, and the generated and genuine Showerthoughts being in the same cluster indicates that the models are successful in replicating the distinct writing of each subreddit (Figure~\ref{figure:subreddits_comparison}, in Appendix).

\subsection{Survey Results}
\label{subsec:survey_results}

A total of 56 human evaluators took our survey (25 participants in Group A and 31 in Group B), resulting in an accumulated 2,520 ratings for the full set of 90 Showerthoughts and an average of 28 ratings per item, as each group reviewed a completely different set of 45 texts.

\paragraph{Demographics of Survey Participants}

The participants' demographics are influenced by the channels the survey was shared in: 
The majority of the participants are younger than 30 years old, with 
8.5\% above 30 years. %
89.4\% of respondents have some degree of machine learning (ML) experience, %
42.6\% have trained an ML model at least 
once, and some of these even work with ML models daily.
Only 10.6\% indicated little to no experience with ML.
53.1\% of participants rarely or never visit Reddit, while the rest visit monthly (8.5\%), weekly (38.3\%), or daily (27.7\%).
31.2\% had never heard of r/Showerthoughts before, while 68.7\% visited the community at least once in the past -- 16.6\% are subscribed and follow it regularly, with 6.2\% even occasionally engaging in the community. 

It is clear that this demographic distribution is not representative for the broader population, but a result of the distribution channels used for the survey: the professional and university networks of the authors. 
From a statistical perspective, this is likely to introduce a bias -- however, we find it highly interesting to study this group of individuals nonetheless, as many are experienced with ML and approximately half are familiar with Reddit, which we hypothesize to potentially improve their abilities.

\begin{table}[htbp]
    \begin{center}
    \scalebox{0.85}{
    \begin{tabular}{l|cccc}
        \toprule
        Source &            Genuine & GPT-2         & GPT-Neo       & ChatGPT       \\
        \midrule
        Score &     \textbf{3.71} &     2.42 & \underline{3.40} & 3.23      \\
        Log.\ Val. &  \textbf{4.20} &     3.10 & \underline{3.96} & 3.55      \\
        Creativity &        \textbf{3.63} &     2.42 &  3.23            & \underline{3.45} \\
        Humour &         \textbf{3.18} &     2.10 &  2.74            & \underline{2.85} \\
        Cleverness &        \textbf{3.41} &     2.19 & \underline{3.15} & 3.07      \\
        \bottomrule
    \end{tabular}
    }
    \caption{Mean score (on a six-point scale) for the Showerthought quality criteria (Log.\ Val. = Logical Validity); best score bold, best model underlined}
    \label{table:survey_general}
    \end{center}
\end{table}

\paragraph{Overview of Showerthought Ratings}

Table~\ref{table:survey_general} displays the average response scores %
for the first five evaluation criteria. %
None of the LLMs is able to beat or match the scores of human-written Showerthoughts, but %
some of them get remarkably close.

Among the models, GPT-Neo achieves the best ratings for general score, logical validity, and cleverness, while ChatGPT (based on GPT-3.5-turbo) performs better on creativity and humour.
It appears that the general ability to write a convincing, logical, and clever Showerthought can be learned in fine-tuning, but more abstract abilities like creativity and humour improve with %
model size.

The smallest model, GPT-2, performs the worst, consistently short of human-written Showerthoughts, %
exhibiting an approximately 30\% worse performance.
GPT-Neo and ChatGPT achieve a much smaller margin with an overall average disparity of 6\% and 7\%, respectively.
The evaluators %
consistently prefer human-written texts -- however, the margins are small %
and this does not necessarily have implications for the task of authorship identification, as we show below.

\begin{table*}[htbp]
    \begin{center}
    \scalebox{0.9}{
    \begin{tabular}{l cc c cc c cc c cc}
        \toprule
        & \multicolumn{1}{c}{Overall} && \multicolumn{2}{c}{Reddit Experience} && \multicolumn{2}{c}{ML Experience} && \multicolumn{2}{c}{Showerthoughts Experience}\\
        \cmidrule(lr){2-2}
        \cmidrule(lr){4-5}
        \cmidrule(lr){7-8}
        \cmidrule(lr){10-11}
        Model & && Yes & No && Yes & No && Yes & No\\
        \midrule
        Genuine & 63.8\,\% && 71.3\,\% & 60.2\,\% && 63.2\,\% & 62.3\,\% && 81.6\,\% & 62.0\,\% \\
        GPT-2   & 73.1\,\% && 71.3\,\% & 72.2\,\% && 74.0\,\% & 74.0\,\% && 60.0\,\% & 72.4\,\% \\
        GPT-Neo & 48.1\,\% && 49.0\,\% & 46.6\,\% && 48.0\,\% & 53.5\,\% && 55.0\,\% & 45.7\,\% \\
        ChatGPT & 46.2\,\% && 43.9\,\% & 45.1\,\% && 46.8\,\% & 44.0\,\% && 42.5\,\% & 44.3\,\% \\
        \midrule
        No.\ Participants &   56 &&    21   &  30   &&  25  &   7  &&  5  &   45  \\
        \bottomrule
    \end{tabular}  
    }
    \caption{Survey participants' accuracy in correctly identifying the Showerthought's source}
    \label{table:survey_accuracy}
    \end{center}
\end{table*}

\begin{table}[h]
    \begin{center}
    \scalebox{0.9}{
    \begin{tabular}{llcccr}
        \toprule
        & & Prec. & Rec. & F1 & Support \\
        \midrule
        \multirow{4}{*}{\rotatebox[origin=c]{90}{GPT-2}}         &    Generated & 0.91 & 1.00 & 0.95 & 2,000 \\
        &    Genuine & 1.00 & 0.90 & 0.95 & 2,000 \\
        &    Accuracy & & & 0.95 & 4,000 \\
        &    Average & 0.96 & 0.95 & 0.95 & 4,000 \\
        \midrule
        \multirow{4}{*}{\rotatebox[origin=c]{90}{GPT-Neo}}         &    Generated & 0.84 & 0.99 & 0.91 & 2,000 \\
        &    Genuine & 0.99 & 0.82 & 0.90 & 2,000 \\
        &    Accuracy & & & 0.90 & 4,000 \\
        &    Average & 0.92 & 0.90 & 0.90 & 4,000 \\
        \midrule
        \multirow{4}{*}{\rotatebox[origin=c]{90}{ChatGPT}}         &    Generated & 0.91 & 0.99 & 0.95 & 400 \\
        &    Genuine & 0.99 & 0.91 & 0.94 & 400 \\
        &    Accuracy & & & 0.95 & 800 \\
        &    Average & 0.95 & 0.95 & 0.95 & 800 \\
        \midrule
        \multirow{4}{*}{\rotatebox[origin=c]{90}{Combined}}         &    Generated & 0.82 & 0.95 & 0.88 & 3,000 \\
        &    Genuine & 0.94 & 0.79 & 0.86 & 3,000 \\
        &    Accuracy & & & 0.87 & 6,000 \\
        &    Average & 0.88 & 0.87 & 0.87 & 6,000 \\
        \bottomrule
    \end{tabular}
    }
    \caption{Precision, Recall, F1, and Support of the RoBERTa models trained for Showerthoughts authorship identification (LLM-specific models and one combined model for all LLMs)}
    \label{table:roberta_results}
    \end{center}
\end{table}

\paragraph{Manual Authorship Identification}

From the survey responses regarding authorship of a text, we consider
answers between 1 and 3 as a vote for AI-generated, and answers between 4 and 6 as a vote for human-written text. %
Table~\ref{table:survey_accuracy} displays the average accuracy of the survey's participants in correctly identifying each Showerthought's source. %
For a more granular evaluation, we additionally display the responses by the participants' experience in Reddit, machine learning (ML), and Showerthoughts.\footnote{The participants were considered `experienced' in one of the given categories if they chose one of the top two answers (e.g., visiting Reddit `Weekly' or `Daily') and `unexperienced' if they chose one of the bottom two answers (e.g., visiting Reddit `Never' or `Rarely').}

We find that the survey participants were not able to consistently identify whether a Showerthought was human-written or AI-generated; Between all human-written (genuine) and GPT-2, GPT-Neo, and ChatGPT generated Showerthoughts %
the survey participants were only able to correctly identify 63.8\%, 73.1\%, 48.1\%, and 46.2\%, respectively. For GPT-Neo and ChatGPT, this is worse than (balanced) random guessing, i.e., a strategy that would choose one of the two classes in 50\% of cases. This indicates that GPT-Neo and ChatGPT already generate Showerthoughts sufficiently convincing to mislead human evaluators.
Experience with Reddit and Showerthoughts improves the participants' ability to identify human-written Showerthoughts, but does not improve their ability to detect AI-generated texts consistently.

To investigate whether evaluators are more accurate with higher confidence, we evaluated high-confidence answers only (i.e., 1\,--\,2 and 4\,--\,6). %
However, detection accuracy did not improve. In these cases GPT-2 was detected with an accuracy of 79.6\%, while there were only small improvements in detecting the other sources. 
The detection accuracy regarding GPT-Neo and ChatGPT remained below the random-guess baseline.
Similar to the overall results, experience with Reddit or Showerthoughts only helped in identifying genuine texts.
This shows that independent of their size GPT-Neo and ChatGPT are able to mislead evaluators with the quality of their generated texts.

\paragraph{Participants' Reasoning for Detecting AI-Generated Texts}

At the end of the survey, participants could add explanations for their %
evaluation.
Within the 42 responses, the primary factors were: illogical statements, common sense, good grammar, lack of humour / depth / creativity, and repetitive word or syntax usage. Endowing machines with commonsense knowledge has been a long-standing goal in AI \cite{TandonVardeDeMelo2017Commonsense}, which LLMs address to a significant degree.
The finding that `good grammar' was frequently mentioned
is noteworthy, as many participants believed that machines excel at grammar while errors indicate human authorship.
These findings are consistent with prior research by~\citet{dugan2022-real}, who identified similar factors as the most commonly cited indicators of AI-generated content.

\subsection{Automated Authorship Identification}
\label{subsec:roberta_results}

This section presents the evaluation results of the four different RoBERTa classifiers introduced in Section \ref{subsec:detection_with_roberta} -- three LLM-specific classifiers and one trained on the combined texts of all three models.
The classification reports presented in Table~\ref{table:roberta_results} show that the classifiers trained per model achieve an overall accuracy ranging from 90\% to 95\%, with the single model trained for all LLMs scoring an accuracy of 87\% (Table~\ref{table:roberta_results}).
Across all classifiers, recall for LLM-generated instances approaches 100\% with lower precision, while precision for the genuine human-authored class is nearly perfect but with lower recall. %
These findings indicate the following:
(1) These classifiers outperform human evaluators on authorship identification.\footnote{%
Note: While human evaluators receive a more general instruction at the beginning of the survey, the classification models are fine-tuned for the task. Nonetheless, we consider this a realistic setup, as almost 70\% of the evaluators have responded to have prior experience with the Showerthoughts community. For future work, human evaluators could be presented with human-written and AI-generated examples at the beginning of the survey.},
(2) The classifiers consistently misclassify a portion of genuine Showerthoughts as generated, which are either lower-quality examples or similar to generated texts in some regard.
(3) The models perform well in detecting the AI-generated texts, with the combined RoBERTa model achieving an average F1 score of 0.87.
(4) Current (GPT-based) language models, independent of their size, appear to utilize similarly transparent techniques for language generation and are therefore similarly easy to detect for an ML classifier, even when trained on a different GPT-based model.

In an additional experiment, we trained a classifier to distinguish texts of GPT-Neo from genuine ones but evaluate its performance on texts of the other LLMs.
The results in Table~\ref{tab:roberta_results_transferred} show that the classifier's average performance on the texts of other models can achieve a relatively high value of 0.86 when a single model's texts are utilized for evaluation.
However, the results are significantly worse when texts of various models, of which most were not part of the training, are included for evaluation, suggesting fine-tuning with texts from multiple LLMs for better detection performance.

Our evaluation of the fine-tuned RoBERTa models shows that
none of the classifiers attain 100\% accuracy, emphasizing caution when using detection tools, particularly in cases with serious consequences such as academic failure or job loss. 
In a real-world setting, the specific LLM invoked to generate and spread texts will likely be unknown, and, therefore, cannot provide training samples, which requires robust generalizable classifiers and non-GPT-based LLMs -- important questions requiring investigation in future work.
Nonetheless, our results suggest that the models have learned patterns that strongly indicate whether a given Showerthought is AI-generated, which proves valuable for evaluating the tokens and patterns that contribute the most to the classification results, which we do in the following section. 

\begin{table}[htbp]
    \begin{center}
    \scalebox{0.9}{
    \begin{tabular}{llcccr}
        \toprule
        & & Prec. & Rec. & F1 & Support \\
        \midrule
        \multirow{4}{*}{\rotatebox[origin=c]{90}{GPT-2}}         &    Generated & 0.83 & 0.90 & 0.86 & 2,000 \\
        &    Genuine & 0.89 & 0.82 & 0.85 & 2,000 \\
        &    Accuracy & & & 0.86 & 4,000 \\
        &    Average & 0.86 & 0.86 & 0.86 & 4,000 \\
        \midrule
        \multirow{4}{*}{\rotatebox[origin=c]{90}{ChatGPT}}         &    Generated & 0.99 & 0.74 & 0.85 & 400 \\
        &    Genuine & 0.79 & 0.99 & 0.88 & 400 \\
        &    Accuracy & & & 0.86 & 800 \\
        &    Average & 0.89 & 0.86 & 0.86 & 800 \\
        \midrule
        \multirow{4}{*}{\rotatebox[origin=c]{90}{All models}}         &    Generated & 0.75 & 0.54 & 0.62 & 3,000 \\
        &    Genuine & 0.64 & 0.82 & 0.72 & 3,000 \\
        &    Accuracy & & & 0.68 & 6,000 \\
        &    Average & 0.69 & 0.68 & 0.67 & 6,000 \\
        \bottomrule
    \end{tabular}
    }
    \caption{Evaluation of the RoBERTa model trained on on GPT-Neo's generated texts when evaluated on texts from other sources}
    \label{tab:roberta_results_transferred}
    \end{center}
\end{table}

\paragraph{Tokens with Greatest Contribution towards Class Prediction}

We use the LLM explainability library {\tt transformers-interpret} %
to identify the most influential tokens per RoBERTa model.
For evaluating correctly and falsely classified texts, we select the top four contributing tokens to each Showerthought's predicted class, then aggregate and normalize each token's significance relative to the dataset. 

The results for the three LLMs are similar -- significant contributors are (1) tokens at the beginning of a sentence, as they start with a capitalized first letter (`If', `The', `You' and `We' seem to be frequent in generated texts) %
and (2) punctuation (`.' and `,' specifically).
Punctuation and specific stop words (e.g., `you', `the') seem to be tokens with high attribution scores for the genuine class, indicating that a critical difference between the two classes %
is the placement of these tokens.
ChatGPT shows slightly different top contributors, especially `Why' and `?' -- this model seems to generate questions more frequently and seems to have a unique usage of the word `is'.
Differences between ChatGPT and the other models may result from ChatGPT's pre-training data including a different subset of Reddit data and the model's much larger size. %

Furthermore, our results indicate that those human-written Showerthoughts falsely classified as AI-generated by GPT-2 and GPT-Neo share the characteristics identified of generated texts, e.g., starting sentences with `You', `The', and `We'.
ChatGPT shows fewer distinct patterns in contributor variety and overlap between correct and incorrect human classifications. 
Showerthoughts mistaken as human-written ones use punctuation and blank spaces in a similar way as the genuine texts, while the misclassified human-written texts use words that may occur rarely, or seem to originate from another language. %
We provide more detailed results in the Appendix.
In summary, RoBERTa classifiers have difficulties in cases where the characteristic writing styles of the classes overlap (especially for GPT-2 and GPT-Neo) or the misclassified Showerthought contains rarely-used or foreign words. %

\section{Conclusion}
In this study, we demonstrate that relatively small, GPT-based LLMs can be fine-tuned to replicate the writing style of short texts of high creative quality, using the Showerthoughts subreddit as an example. While it remains to be investigated to what extent the creativity stems from observations encountered in the pretraining corpus as opposed to novel creations, we find that large numbers of diverse texts can be produced with great ease.
Human raters confirm that the generated texts exhibit wit, creativity, and humour. 
This paves the way for diverse applications in productivity, creative work, and entertainment, and is relevant for practitioners deploying small LLMs to be cost-efficient.

We find that human evaluators rate the generated texts on average slightly lower regarding creativity, humour, cleverness. 
This does not seem to aid in authorship detection (“I believe this Showerthought has been written by a real person”), as we find that evaluators could not reliably distinguish AI-generated texts from human-written ones. Additionally, the quality of human-written Showerthoughts varies, with bad ones often being mislabeled as AI-generated.

Nonetheless, the possibility to abuse these models to produce spam, misinformation, or other harmful content is a growing concern.
Our RoBERTa-based authorship identification classifiers performs well after fine-tuning, revealing interesting hidden patterns that help in detecting the texts generated by specific LLMs.
While ML classifiers can currently detect AI-generated texts (when fine-tuned for the task), we can assume that the text generation quality of LLMs will further improve, making this task more difficult.
Additionally, differently designed models may pursue other strategies for generating texts, necessitating their inclusion when training general-purpose classifiers.

Our work extends existing work that LLMs can learn to generate specific types of texts (when fine-tuned on high-quality data) to the domain of creative and witty texts, as exhibited by Showerthoughts, but not limited to those.
For example, practitioners who would like to utilize such a LLM for marketing or copy-writing, could not only prompt it for general Showerthoughts about a random topic, but also add the start of a text or topic to their prompt for the LLM to complete. 
Alternatively, generated texts can be clustered by topic to identify the right topics for a specific use case.
Simultaneously, we strongly recommend further research on detection mechanisms -- while training detection models using generated texts of known LLMs and those fine-tuned on known datasets seems feasible, the task becomes more difficult when there is an exceedingly high number of LLMs to consider and even more so if the author-LLM's architecture or the training dataset is not known.

\section*{Ethics Statement}
As the dataset proposed in this paper (see Section~\ref{subsec:dataset}) is based on real user-submitted data from the Reddit Showerthoughts community, it is important to handle it with care. It should not be used to identify individuals and might contain offensive text or wrong information. This should be considered in future use of the dataset.
For the survey (see Section~\ref{subsec:survey}), we manually removed inappropriate content to make it appropriate for the context of where the survey was distributed, e.g., university mailing lists. 
The type of survey conducted here is exempt from an ethics board review at our institution, as we have carefully designed it to be transparently described and to avoid collection of personal data. 

\bibliography{references}

\begin{thebibliography}{41}
\expandafter\ifx\csname natexlab\endcsname\relax\def\natexlab#1{#1}\fi

\bibitem[{Abokhodair et~al.(2015)Abokhodair, Yoo, and McDonald}]{abokhodair2015dissecting}
Norah Abokhodair, Daisy Yoo, and David~W. McDonald. 2015.
\newblock \href {https://doi.org/10.1145/2675133.2675208} {{Dissecting a Social Botnet: Growth, Content and Influence in Twitter}}.
\newblock In \emph{Proceedings of the 18th ACM Conference on Computer Supported Cooperative Work \& Social Computing}, CSCW '15, page 839–851, New York, NY, USA. Association for Computing Machinery.

\bibitem[{Agrawal et~al.(2022)Agrawal, Buz, and de~Melo}]{AgrawalBuzDeMelo2022WallStreetBetsSocial}
Pratik Agrawal, Tolga Buz, and Gerard de~Melo. 2022.
\newblock \href {https://aisel.aisnet.org/amcis2022/sig\_sc/sig\_sc/8} {{WallStreetBets} beyond {GameStop}, {YOLOs}, and the {Moon}: {The} unique traits of {Reddit}'s finance communities}.
\newblock In \emph{Proceedings of AMCIS 2022}. Association for Information Systems.

\bibitem[{Alada{\u{g}} et~al.(2018)Alada{\u{g}}, Muderrisoglu, Akbas, Zahmacioglu, and Bingol}]{aladaug2018detecting}
Ahmet~Emre Alada{\u{g}}, Serra Muderrisoglu, Naz~Berfu Akbas, Oguzhan Zahmacioglu, and Haluk~O Bingol. 2018.
\newblock {Detecting Suicidal Ideation on Forums: Proof-of-Concept Study}.
\newblock \emph{Journal of Medical Internet Research}, 20(6):e9840.

\bibitem[{Brown et~al.(2020)Brown, Mann, Ryder, Subbiah, Kaplan, Dhariwal, Neelakantan, Shyam, Sastry, Askell, Agarwal, Herbert-Voss, Krueger, Henighan, Child, Ramesh, Ziegler, Wu, Winter, Hesse, Chen, Sigler, Litwin, Gray, Chess, Clark, Berner, McCandlish, Radford, Sutskever, and Amodei}]{brown2020-language}
Tom Brown, Benjamin Mann, Nick Ryder, Melanie Subbiah, Jared~D Kaplan, Prafulla Dhariwal, Arvind Neelakantan, Pranav Shyam, Girish Sastry, Amanda Askell, Sandhini Agarwal, Ariel Herbert-Voss, Gretchen Krueger, Tom Henighan, Rewon Child, Aditya Ramesh, Daniel Ziegler, Jeffrey Wu, Clemens Winter, Chris Hesse, Mark Chen, Eric Sigler, Mateusz Litwin, Scott Gray, Benjamin Chess, Jack Clark, Christopher Berner, Sam McCandlish, Alec Radford, Ilya Sutskever, and Dario Amodei. 2020.
\newblock \href {https://proceedings.neurips.cc/paper/2020/file/1457c0d6bfcb4967418bfb8ac142f64a-Paper.pdf} {{Language Models are Few-Shot Learners}}.
\newblock In \emph{Advances in Neural Information Processing Systems}, volume~33, pages 1877--1901. Curran Associates, Inc.

\bibitem[{Buz et~al.(2024)Buz, Schneider, Kaffee, and de~Melo}]{BuzSchneiderKaffeeDeMelo2024WSBPredictions}
Tolga Buz, Moritz Schneider, Lucie-Aimée Kaffee, and Gerard de~Melo. 2024.
\newblock \href {https://doi.org/https://doi.org/10.1145/3614419.3643993} {{Highly Regarded Investors? Mining Predictive Value from the Collective Intelligence of Reddit’s WallStreetBets}}.
\newblock In \emph{ACM Web Science Conference (Websci ’24), May 21--24, 2024, Stuttgart, Germany}, page~11. ACM, New York, NY, USA.

\bibitem[{Chakrabarty et~al.(2022{\natexlab{a}})Chakrabarty, Choi, and Shwartz}]{chakrabarty2022s}
Tuhin Chakrabarty, Yejin Choi, and Vered Shwartz. 2022{\natexlab{a}}.
\newblock {It’s not Rocket Science: Interpreting Figurative Language in Narratives}.
\newblock \emph{Transactions of the Association for Computational Linguistics}, 10:589--606.

\bibitem[{Chakrabarty et~al.(2023{\natexlab{a}})Chakrabarty, Laban, Agarwal, Muresan, and Wu}]{DBLP:journals/corr/abs-2309-14556}
Tuhin Chakrabarty, Philippe Laban, Divyansh Agarwal, Smaranda Muresan, and Chien{-}Sheng Wu. 2023{\natexlab{a}}.
\newblock \href {https://doi.org/10.48550/arXiv.2309.14556} {{Art or Artifice? {L}arge Language Models and the False Promise of Creativity}}.
\newblock \emph{CoRR}, abs/2309.14556.

\bibitem[{Chakrabarty et~al.(2023{\natexlab{b}})Chakrabarty, Padmakumar, Brahman, and Muresan}]{chakrabarty2023creativity}
Tuhin Chakrabarty, Vishakh Padmakumar, Faeze Brahman, and Smaranda Muresan. 2023{\natexlab{b}}.
\newblock {Creativity Support in the Age of Large Language Models: An Empirical Study Involving Emerging Writers}.
\newblock \emph{arXiv preprint arXiv:2309.12570}.

\bibitem[{Chakrabarty et~al.(2022{\natexlab{b}})Chakrabarty, Padmakumar, and He}]{chakrabarty2022help}
Tuhin Chakrabarty, Vishakh Padmakumar, and He~He. 2022{\natexlab{b}}.
\newblock {Help me write a poem: Instruction Tuning as a Vehicle for Collaborative Poetry Writing}.
\newblock \emph{arXiv preprint arXiv:2210.13669}.

\bibitem[{Chen and Shu(2024)}]{chen2024llmgenerated}
Canyu Chen and Kai Shu. 2024.
\newblock \href {http://arxiv.org/abs/2309.13788} {{Can LLM-Generated Misinformation Be Detected?}}
\newblock \emph{arXiv preprint arXiv:2309.13788}.

\bibitem[{Clark et~al.(2021)Clark, August, Serrano, Haduong, Gururangan, and Smith}]{clark2021-all}
Elizabeth Clark, Tal August, Sofia Serrano, Nikita Haduong, Suchin Gururangan, and Noah~A. Smith. 2021.
\newblock \href {https://doi.org/10.18653/v1/2021.acl-long.565} {{"All That{'}s {`}Human{'} Is Not Gold: Evaluating Human Evaluation of Generated Text"}}.
\newblock In \emph{Proceedings of the 59th Annual Meeting of the Association for Computational Linguistics and the 11th International Joint Conference on Natural Language Processing (Volume 1: Long Papers)}, pages 7282--7296, Online. Association for Computational Linguistics.

\bibitem[{Crawford(2020)}]{crawford2020daydreaming}
Kevin Crawford. 2020.
\newblock {Daydreaming of Genius: Insight and the Wandering Mind}.
\newblock \emph{Scientific Kenyon: The Neuroscience Edition}, 4(1):55--62.

\bibitem[{Davis et~al.(2024)Davis, Van~Bulck, Durieux, and Lindvall}]{davis2024temperature}
Joshua Davis, Liesbet Van~Bulck, Brigitte~N Durieux, and Charlotta Lindvall. 2024.
\newblock \href {https://doi.org/10.2196/53559} {{"The Temperature Feature of ChatGPT: Modifying Creativity for Clinical Research"}}.
\newblock \emph{JMIR Hum Factors}, 11.

\bibitem[{Deng et~al.(2023)Deng, Gao, Miao, and Zhang}]{deng2023efficient}
Zhijie Deng, Hongcheng Gao, Yibo Miao, and Hao Zhang. 2023.
\newblock {Efficient Detection of LLM-generated Texts with a Bayesian Surrogate Model}.
\newblock \emph{arXiv preprint arXiv:2305.16617}.

\bibitem[{Dugan et~al.(2022)Dugan, Ippolito, Kirubarajan, Shi, and Callison-Burch}]{dugan2022-real}
Liam Dugan, Daphne Ippolito, Arun Kirubarajan, Sherry Shi, and Chris Callison-Burch. 2022.
\newblock {Real or Fake Text?: Investigating Human Ability to Detect Boundaries Between Human-Written and Machine-Generated Text}.
\newblock \emph{arXiv preprint arXiv:2212.12672}.

\bibitem[{Dwivedi et~al.(2023)Dwivedi, Kshetri, Hughes, Slade, Jeyaraj, Kar, Baabdullah, Koohang, Raghavan, Ahuja et~al.}]{dwivedi2023so}
Yogesh~K Dwivedi, Nir Kshetri, Laurie Hughes, Emma~Louise Slade, Anand Jeyaraj, Arpan~Kumar Kar, Abdullah~M Baabdullah, Alex Koohang, Vishnupriya Raghavan, Manju Ahuja, et~al. 2023.
\newblock {“So what if ChatGPT wrote it?” Multidisciplinary perspectives on opportunities, challenges and implications of generative conversational AI for research, practice and policy}.
\newblock \emph{International Journal of Information Management}, 71:102642.

\bibitem[{Gao et~al.(2020)Gao, Biderman, Black, Golding, Hoppe, Foster, Phang, He, Thite, Nabeshima et~al.}]{gao2020pile}
Leo Gao, Stella Biderman, Sid Black, Laurence Golding, Travis Hoppe, Charles Foster, Jason Phang, Horace He, Anish Thite, Noa Nabeshima, et~al. 2020.
\newblock {The Pile: An 800GB Dataset of Diverse Text for Language Modeling}.
\newblock \emph{arXiv preprint arXiv:2101.00027}.

\bibitem[{Gerv{\'a}s(2009)}]{gervas2009computational}
Pablo Gerv{\'a}s. 2009.
\newblock {Computational Approaches to Storytelling and Creativity}.
\newblock \emph{AI Magazine}, 30(3):49--49.

\bibitem[{Ippolito et~al.(2020)Ippolito, Duckworth, Callison-Burch, and Eck}]{ippolito2020-automatic}
Daphne Ippolito, Daniel Duckworth, Chris Callison-Burch, and Douglas Eck. 2020.
\newblock \href {https://doi.org/10.18653/v1/2020.acl-main.164} {{Automatic Detection of Generated Text is Easiest when Humans are Fooled}}.
\newblock In \emph{Proceedings of the 58th Annual Meeting of the Association for Computational Linguistics}, pages 1808--1822, Online. Association for Computational Linguistics.

\bibitem[{Jentzsch and Kersting(2023)}]{jentzsch2023chatgpt}
Sophie~F. Jentzsch and Kristian Kersting. 2023.
\newblock \href {https://doi.org/10.18653/v1/2023.wassa-1.29} {{ChatGPT is fun, but it is not funny! Humor is still challenging Large Language Models}}.
\newblock pages 325--340.

\bibitem[{Kenny et~al.(2022)Kenny, Fischhoff, Davis, Carley, and Canfield}]{kenny2022duped}
Ryan Kenny, Baruch Fischhoff, Alex Davis, Kathleen~M Carley, and Casey Canfield. 2022.
\newblock {Duped by Bots: Why Some are Better than Others at Detecting Fake Social Media Personas}.
\newblock \emph{Human factors}, page 00187208211072642.

\bibitem[{Kim et~al.(2023)Kim, Han, and Choi}]{kim2023predicting}
Jinhyeon Kim, Jinyoung Han, and Daejin Choi. 2023.
\newblock {Predicting continuity of online conversations on Reddit}.
\newblock \emph{Telematics and Informatics}, 79:101965.

\bibitem[{Kincaid et~al.(1975)Kincaid, Fishburne~Jr, Rogers, and Chissom}]{kincaid1975derivation}
J~Peter Kincaid, Robert~P Fishburne~Jr, Richard~L Rogers, and Brad~S Chissom. 1975.
\newblock Derivation of new readability formulas (automated readability index, fog count and flesch reading ease formula) for navy enlisted personnel.

\bibitem[{Kreps et~al.(2022)Kreps, McCain, and Brundage}]{kreps2022-all}
Sarah Kreps, R.~Miles McCain, and Miles Brundage. 2022.
\newblock \href {https://doi.org/10.1017/XPS.2020.37} {{All the News That’s Fit to Fabricate: AI-Generated Text as a Tool of Media Misinformation}}.
\newblock \emph{Journal of Experimental Political Science}, 9(1):104–117.

\bibitem[{Kumar and Shah(2018)}]{kumar2018false}
Srijan Kumar and Neil Shah. 2018.
\newblock {False Information on Web and Social Media: A Survey}.
\newblock \emph{arXiv preprint arXiv:1804.08559}.

\bibitem[{Köbis and Mossink(2021)}]{kobis2021-artificial}
Nils Köbis and Luca~D. Mossink. 2021.
\newblock \href {https://doi.org/https://doi.org/10.1016/j.chb.2020.106553} {{Artificial intelligence versus Maya Angelou: Experimental evidence that people cannot differentiate AI-generated from human-written poetry}}.
\newblock \emph{Computers in Human Behavior}, 114:106553.

\bibitem[{Lin et~al.(2017)Lin, Salehi, Yao, Chen, and Bernstein}]{lin2017better}
Zhiyuan Lin, Niloufar Salehi, Bowen Yao, Yiqi Chen, and Michael Bernstein. 2017.
\newblock {Better When It Was Smaller? Community Content and Behavior After Massive Growth}.
\newblock In \emph{Proceedings of the International AAAI Conference on Web and Social Media}, volume~11, pages 132--141.

\bibitem[{Liu et~al.(2019)Liu, Ott, Goyal, Du, Joshi, Chen, Levy, Lewis, Zettlemoyer, and Stoyanov}]{liu2019roberta}
Yinhan Liu, Myle Ott, Naman Goyal, Jingfei Du, Mandar Joshi, Danqi Chen, Omer Levy, Mike Lewis, Luke Zettlemoyer, and Veselin Stoyanov. 2019.
\newblock {RoBERTa: A Robustly Optimized BERT Pretraining Approach}.
\newblock \emph{arXiv preprint arXiv:1907.11692}.

\bibitem[{Manduchi et~al.(2024)Manduchi, Pandey, Bamler, Cotterell, Däubener, Fellenz, Fischer, Gärtner, Kirchler, Kloft, Li, Lippert, de~Melo, Nalisnick, Ommer, Ranganath, Rudolph, Ullrich, den Broeck, Vogt, Wang, Wenzel, Wood, Mandt, and Fortuin}]{ManduchiEtAl2024GenAI}
Laura Manduchi, Kushagra Pandey, Robert Bamler, Ryan Cotterell, Sina Däubener, Sophie Fellenz, Asja Fischer, Thomas Gärtner, Matthias Kirchler, Marius Kloft, Yingzhen Li, Christoph Lippert, Gerard de~Melo, Eric Nalisnick, Björn Ommer, Rajesh Ranganath, Maja Rudolph, Karen Ullrich, Guy~Van den Broeck, Julia~E Vogt, Yixin Wang, Florian Wenzel, Frank Wood, Stephan Mandt, and Vincent Fortuin. 2024.
\newblock \href {http://arxiv.org/abs/2403.00025} {On the challenges and opportunities in generative ai}.

\bibitem[{Mitchell et~al.(2023)Mitchell, Lee, Khazatsky, Manning, and Finn}]{mitchell2023-detectgpt}
Eric Mitchell, Yoonho Lee, Alexander Khazatsky, Christopher~D Manning, and Chelsea Finn. 2023.
\newblock {DetectGPT: Zero-Shot Machine-Generated Text Detection using Probability Curvature}.
\newblock \emph{arXiv preprint arXiv:2301.11305}.

\bibitem[{Muñoz-Ortiz et~al.(2023)Muñoz-Ortiz, Gómez-Rodríguez, and Vilares}]{munozortiz2023contrasting}
Alberto Muñoz-Ortiz, Carlos Gómez-Rodríguez, and David Vilares. 2023.
\newblock \href {http://arxiv.org/abs/2308.09067} {{Contrasting Linguistic Patterns in Human and LLM-Generated Text}}.

\bibitem[{Noy and Zhang(2023)}]{noy2023experimental}
Shakked Noy and Whitney Zhang. 2023.
\newblock \href {https://doi.org/10.1126/science.adh2586} {Experimental evidence on the productivity effects of generative artificial intelligence}.
\newblock \emph{Science}, 381(6654):187--192.

\bibitem[{Roemmele(2021)}]{roemmele2021inspiration}
Melissa Roemmele. 2021.
\newblock \href {http://arxiv.org/abs/2107.04007} {{Inspiration through Observation: Demonstrating the Influence of Automatically Generated Text on Creative Writing}}.

\bibitem[{Sejnowski(2023)}]{sejnowski2023large}
Terrence~J Sejnowski. 2023.
\newblock {Large Language Models and the Reverse Turing Test}.
\newblock \emph{Neural computation}, 35(3):309--342.

\bibitem[{Shao et~al.(2018)Shao, Ciampaglia, Varol, Yang, Flammini, and Menczer}]{shao2018spread}
Chengcheng Shao, Giovanni~Luca Ciampaglia, Onur Varol, Kai-Cheng Yang, Alessandro Flammini, and Filippo Menczer. 2018.
\newblock The spread of low-credibility content by social bots.
\newblock \emph{Nature communications}, 9(1):1--9.

\bibitem[{Shazeer and Stern(2018)}]{shazeer2018adafactor}
Noam Shazeer and Mitchell Stern. 2018.
\newblock {Adafactor: Adaptive Learning Rates with Sublinear Memory Cost}.
\newblock In \emph{International Conference on Machine Learning}, pages 4596--4604. PMLR.

\bibitem[{Tandon et~al.(2017)Tandon, Varde, and {de Melo}}]{TandonVardeDeMelo2017Commonsense}
Niket Tandon, Aparna Varde, and Gerard {de Melo}. 2017.
\newblock \href {https://dl.acm.org/citation.cfm?id=3186562} {Commonsense knowledge in machine intelligence}.
\newblock \emph{SIGMOD Record}, 46(4):49--52.

\bibitem[{Tang et~al.(2024)Tang, Chuang, and Hu}]{tang2024science}
Ruixiang Tang, Yu-Neng Chuang, and Xia Hu. 2024.
\newblock {The Science of Detecting LLM-Generated Text}.
\newblock \emph{Communications of the ACM}, 67(4):50--59.

\bibitem[{Tang et~al.(2021)Tang, Fabbri, Li, Mao, Adams, Wang, Celikyilmaz, Mehdad, and Radev}]{tang2021investigating}
Xiangru Tang, Alexander Fabbri, Haoran Li, Ziming Mao, Griffin~Thomas Adams, Borui Wang, Asli Celikyilmaz, Yashar Mehdad, and Dragomir Radev. 2021.
\newblock {Investigating Crowdsourcing Protocols for Evaluating the Factual Consistency of Summaries}.
\newblock \emph{arXiv preprint arXiv:2109.09195}.

\bibitem[{Van~der Maaten and Hinton(2008)}]{van2008visualizing}
Laurens Van~der Maaten and Geoffrey Hinton. 2008.
\newblock {Visualizing data using t-SNE}.
\newblock \emph{Journal of machine learning research}, 9(11).

\bibitem[{Wani and Jabin(2017)}]{wani2017sneak}
Mudasir~Ahmad Wani and Suraiya Jabin. 2017.
\newblock {A sneak into the Devil's Colony-Fake Profiles in Online Social Networks}.
\newblock \emph{arXiv preprint arXiv:1705.09929}.

\end{thebibliography}

\appendix

\section{Appendix}
\label{sec:appendix}

\subsection{Simulation of Human Preference with GPT-4}

We conducted an additional experiment using OpenAI's GPT-4 API investigating its ability to learn from the survey results to simulate the preferences of the human evaluators on a larger set of Showerthoughts. 
For this purpose, we defined a system prompt that includes a set of survey-evaluated examples and their average scores for all six categories to provide guidance for the model. 
To measure whether the few-shot prompting has a genuine effect and how the number of few-shot examples affects the results, we experimented with different amounts of examples, starting with three (3\% of all survey items), 45 (50\%), and 72 (80\%), while using the rest of the survey items for testing. 

We measured the coherence of GPT-4's test outputs with the Pearson Correlation metric, which shows a significant increase in correlation when increasing the number of examples shown to GPT-4 in the system prompt: 
after three examples (train), GPT-4's ratings obtain a Pearson correlation of 0.28 with the remaining human evaluations (test), whereas the correlation is 0.49 after 45 examples (50\%), and 0.70 after 72 examples (which is a 80--20 train--test split). In order to further validate these results, we perform tenfold cross-validation using all 90 evaluated Showerthoughts, i.e., by splitting up the evaluated examples into groups of nine and using each group as a test set in a separate iteration, while all other groups are shown to the model as few-shot examples.

The system prompt is defined as follows:

\begin{quote}
   \textit{Act like a frequent visitor of Reddit, and its r/Showerthoughts community in particular. You participate in a scientific survey and utilize your experience to rate Showerthoughts across five dimensions: general score, validity, creativity, funniness, cleverness - with scores from 1 (low) to 6 (high). Additionally, you make a guess on a range from 1 to 6 whether the Showerthought was written by a human author (6) or generated by a language model (1). In order to learn how to score the Showerthoughts, you will receive examples, which have been rated by a team of human annotators. Your task is to rate Showerthoughts as similar to the human annotators as possible.}
    
   \textit{Here are the examples:} \\
   \textit{1	Most drivers of the Honda Fit are in fact not fit 	3,8	3,4	4	4,28	3,2	4,28 } \\
   \dots
\end{quote}

\noindent For the evaluation of a larger set of 2,000 Showerthoughts per source, we provide all human-labelled items as examples within the system prompt to maximize the model's ability to simulate human preference.

\subsection{Visualization of Sentence Embeddings}

\begin{figure}[htbp]
  \centering
  \includegraphics[width=\linewidth]{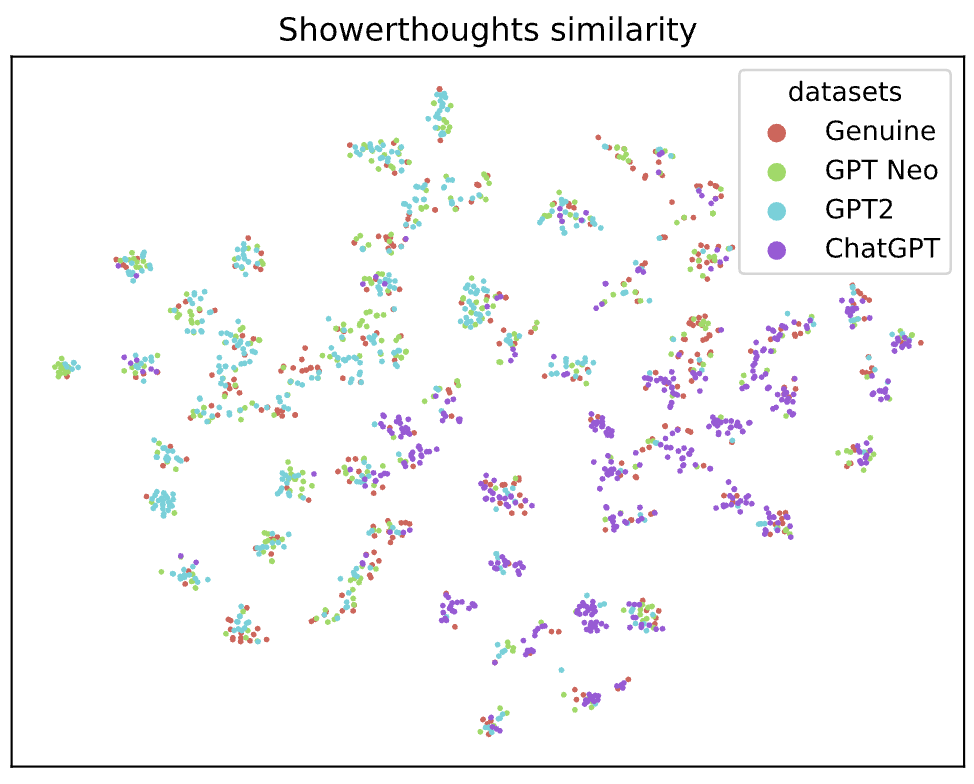}
  \caption{\centering Semantic diversity of the different Showerthoughts datasets (t-SNE visualization of SBERT embeddings)}
  \label{figure:embeddings_comparison}
\end{figure}

\begin{figure}[htbp]
  \centering
  \includegraphics[width=\linewidth]{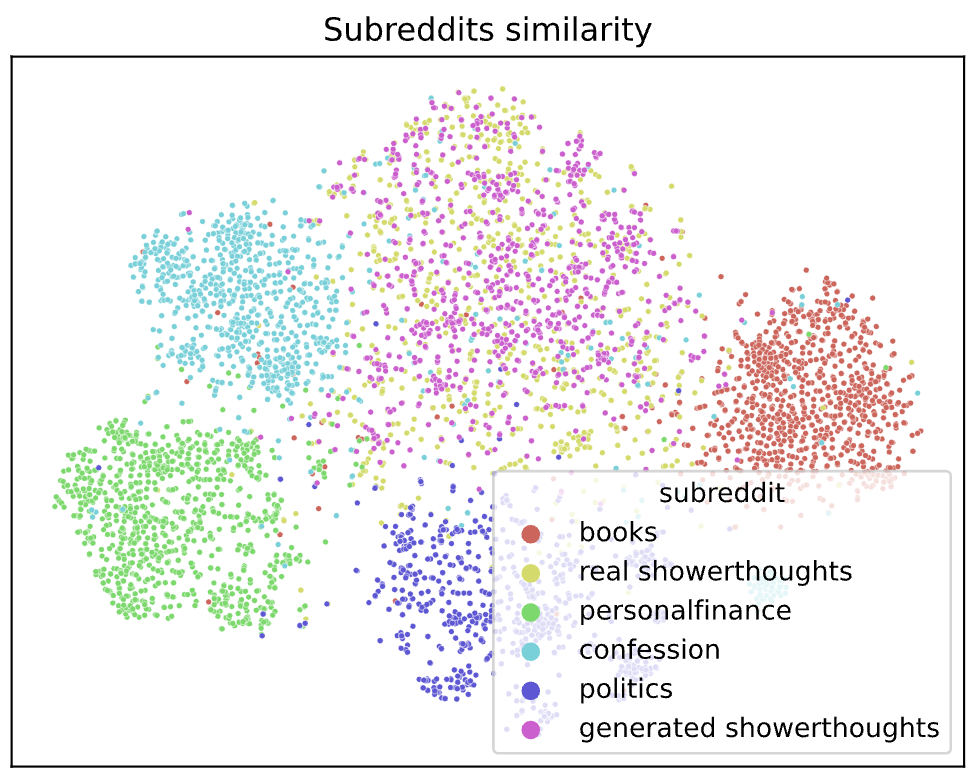}
  \caption{\centering Comparison of genuine and generated Showerthoughts embeddings to other relevant subreddits}
  \label{figure:subreddits_comparison}
\end{figure}

Figures \ref{figure:embeddings_comparison} and \ref{figure:subreddits_comparison} show the distribution of the SBERT embeddings for the different Showerthoughts compared to each other and compared to different subreddits selected for their similarity, respectively.
These indicate that the fine-tuned LLMs in fact reproduce all topics that the original Showerthoughts cover, while ChatGPT is limited to a subset of the topics.

\subsection{Further Survey Details}
\label{appendix:survey}

This section provides additional information on the conducted survey.%

\subsubsection{Participant Briefing}

All survey participants were briefed with the following text:

``We are a group of students from the Hasso Plattner Institute in Potsdam who are taking part in the research seminar `Recent Trends in AI and Deep Learning'. 
As part of the project, we have trained a Machine Learning model that is able to generate short texts in the style of the Reddit community `Showerthoughts'. This survey aims to evaluate the quality of the generated texts compared to the original examples.''

\paragraph{Definition}
\begin{itemize}
    \item \textbf{Reddit} is a social media platform that is organized in sub-communities called ``subreddits''. Any user can create a subreddit that revolves around any specific topic, e.g., world news, formula 1, a specific computer game, or the newest Apple iPhone. Users interested in a community can subscribe and interact within the community by posting content (self-written texts, images, videos, or links to external websites), commenting on posts, or up-/downvoting other posts and comments. Each subreddit usually has a self-defined set of rules and guidelines and is managed by a group of moderators.
    \item The community of \textbf{r/showerthoughts} describes itself as a ``subreddit for sharing those miniature epiphanies you have that highlight the oddities within the familiar.'' They define a ``Showerthought'' as ``a loose term that applies to the type of thoughts you might have while carrying out a routine task like showering, driving, or daydreaming. At their best, showerthoughts are universally relatable and find the amusing/interesting within the mundane.''
\end{itemize}

\paragraph{Survey Setup}
After a few demographic questions, you will be presented with Showerthoughts, of which some are real examples from the community, and some are generated by one of three Machine Learning models (GPT-2, GPT Neo, and ChatGPT).
The survey results will be anonymised and utilised only in this research project and the resulting paper.
This survey consists of 5 demographics-related questions, followed by the 45 Showerthoughts, which have to be rated regarding a set of criteria each. Finally, you can optionally describe what your thinking process was like / what criteria you used to distinguish genuine from generated Showerthoughts.
We estimate that the survey will take you between 20 and 30 minutes to complete.

\subsubsection{Survey Questions}

\begin{table*}[]
    \centering
    \begin{tabular}{p{0.5\linewidth}p{0.5\linewidth}}
    \toprule
        Question & Answer Options \\
        \midrule
        How old are you? & <20 / 20--25 / 26--30 / 31--40/ 41--55/ >55 \\
        How often do you visit Reddit? & Never / Rarely / Monthly / Weekly / Daily \\
        Are you familiar with the r/Showerthoughts community? & No never heard of it / Visited sometime in the past / Subscribed and regularly following / Interact (post, up/downvote, or comment) rarely / Interact (post, up/downvote, or comment) regularly \\
        How experienced are you in using Machine Learning models? & No experience / Using a product with AI or Machine Learning-based features / Played around with AI tools (e.g., ChatGPT) / Trained a ML model at least once / Working with ML models regularly \\
        \bottomrule
    \end{tabular}
    \caption{Demographic Survey Questions and Answer Options}
    \label{tab:demographic_questions}
\end{table*}

After the demographic questions shown in Table \ref{tab:demographic_questions}, the participants were presented a list of 45 Showerthoughts, each with six questions to answer on a six-step Likert scale (from 1= Strongly disagree to 6 = Strongly agree):

\begin{enumerate}
    \item I like this Showerthought.
    \item It makes a true/valid/logical statement.
    \item It is creative.
    \item It is funny.
    \item It is clever.
    \item I believe this Showerthought has been written by a real person.
\end{enumerate}

At the end of the survey, we asked the participants a final optional question that could be answered with a free text:
``When you tried to distinguish genuine from generated Showerthoughts, was there anything specific (e.g., bad grammar, or logical errors) that unveiled the generated ones?''

\subsubsection{Retrieving Random Genuine Showerthoughts}

In order to retrieve random genuine Showerthoughts we used an endpoint Reddit provides with its API\footnote{\url{www.reddit.com/dev/api/\#GET_random}}. To retrieve Showerthoughts specifically we used {\tt GET /r/Showerthoughts/random}.

\subsubsection{Statistics of Demographic Question Results}

\begin{figure}[h]
    \centering
    \begin{tikzpicture}[every label/.append style={font=\scriptsize}]
    \pie[radius=1.6, /tikz/nodes={font=\scriptsize}]{
        8.5/No experience,
        46.8/Played around with AI Tools,
        2.1/Using AI or ML products,
        27.7/Trained a ML model,
        14.9/Working in the ML field
    }
    \end{tikzpicture}
    \caption{Experience in ML}
    \label{fig:experience}
\end{figure}
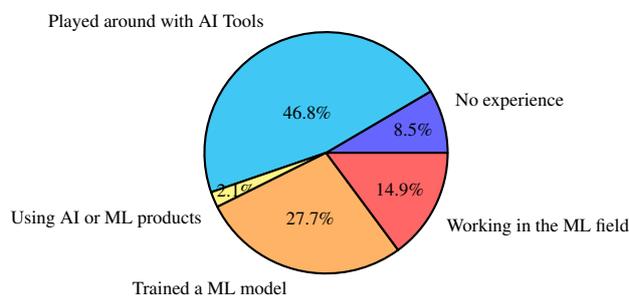

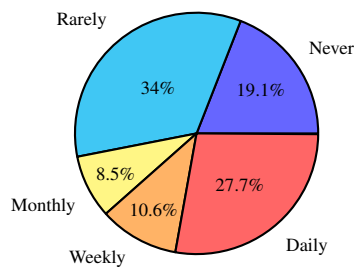
\begin{figure}[h]
  \centering
    \centering
    \begin{tikzpicture}[every label/.append style={font=\scriptsize}]
    \pie[radius=1.6, /tikz/nodes={font=\scriptsize}]{
        19.1/Never,
        34/Rarely,
        8.5/Monthly,
        10.6/Weekly,
        27.7/Daily
    }
    \end{tikzpicture}
    \caption{Reddit usage}
    \label{fig:usage}
\end{figure}

\begin{figure}[h]
  \begin{minipage}{0.45\textwidth}
    \centering
    \begin{tikzpicture}[every label/.append style={font=\scriptsize}]
    \pie[radius=1.6, /tikz/nodes={font=\scriptsize}]{
        31.2/Never heard of it,
        52.1/Visited in the past,
        6.2/Interacting rarely,
        10.4/Regularly following
    }
    \end{tikzpicture}
    \caption{Familarity with Showerthoughts}
    \label{fig:familarity}
  \end{minipage}
\end{figure}
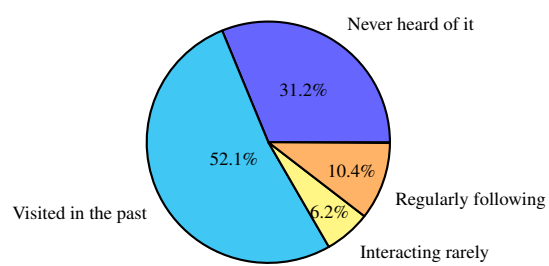

Figures \ref{fig:age} -- \ref{fig:familarity} illustrate the survey participants' demographics and levels of familiarity with machine learning, Reddit, and the Showerthoughts subreddit.

\begin{figure}[h]
  \centering
    \centering
    \begin{tikzpicture}[every label/.append style={font=\scriptsize}]
    \pie[radius=1.6, /tikz/nodes={font=\scriptsize}]{
        17/<20,
        55.3/20--25,
        19.1/26--30,
        6.4/31--40,
        2.1/41--55
    }
    \end{tikzpicture}
    \caption{Age}
    \label{fig:age}
\end{figure}

\subsubsection{Statistics on Showerthought Ratings}

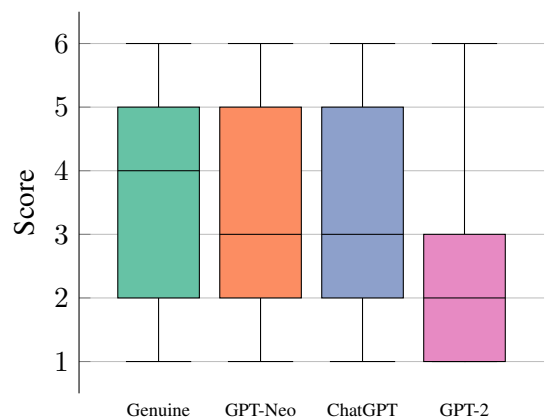
\begin{figure}[htbp]
  \centering
    \centering
    \begin{tikzpicture}
        \begin{axis}[
            width=\linewidth,
            cycle list/Set2-8,
            boxplot/draw direction = y,
            x axis line style = {opacity=0},
            axis x line* = bottom,
            axis y line* = left,
            ymajorgrids,
            xtick = {1, 2, 3, 4},
            xticklabel style = {align=center, font=\scriptsize},
            xticklabels = {Genuine, GPT-Neo, ChatGPT, GPT-2},
            xtick style = {draw=none}, %
            ylabel = {Score},
            ytick = {1, 2, 3, 4, 5, 6}
        ]
        \addplot+[boxplot prepared={
            median=4,
            upper quartile=5,
            lower quartile=2,
            upper whisker=6,
            lower whisker=1
        }, fill, draw=black] coordinates {};
        \addplot+[boxplot prepared={
            median=3,
            upper quartile=5,
            lower quartile=2,
            upper whisker=6,
            lower whisker=1
        }, fill, draw=black] coordinates {};
        \addplot+[boxplot prepared={
            median=3,
            upper quartile=5,
            lower quartile=2,
            upper whisker=6,
            lower whisker=1
        }, fill, draw=black] coordinates {};
        \addplot+[boxplot prepared={
            median=2,
            upper quartile=3,
            lower quartile=1,
            upper whisker=6,
            lower whisker=1
        }, fill, draw=black] coordinates {};
        \end{axis}
    \end{tikzpicture}
    \caption{General Score}
    \label{fig:general_score}
\end{figure}

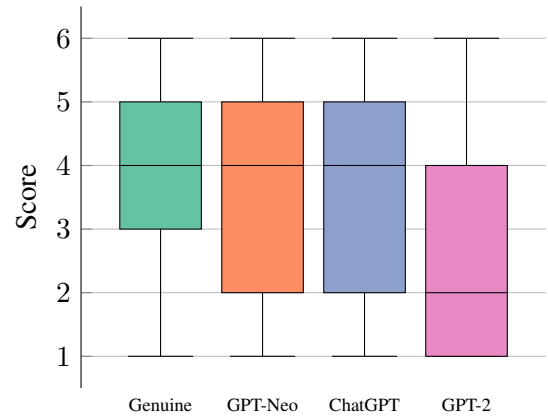
\begin{figure}
    \centering
    \begin{tikzpicture}
        \begin{axis}[
            width=\linewidth,
            cycle list/Set2-8,
            boxplot/draw direction = y,
            x axis line style = {opacity=0},
            axis x line* = bottom,
            axis y line* = left,
            ymajorgrids,
            xtick = {1, 2, 3, 4},
            xticklabel style = {align=center, font=\scriptsize},
            xticklabels = {Genuine, GPT-Neo, ChatGPT, GPT-2},
            xtick style = {draw=none}, %
            ylabel = {Score},
            ytick = {1, 2, 3, 4, 5, 6}
        ]
        \addplot+[boxplot prepared={
            median=4,
            upper quartile=5,
            lower quartile=3,
            upper whisker=6,
            lower whisker=1
        }, fill, draw=black] coordinates {};
        \addplot+[boxplot prepared={
            median=4,
            upper quartile=5,
            lower quartile=2,
            upper whisker=6,
            lower whisker=1
        }, fill, draw=black] coordinates {};
        \addplot+[boxplot prepared={
            median=4,
            upper quartile=5,
            lower quartile=2,
            upper whisker=6,
            lower whisker=1
        }, fill, draw=black] coordinates {};
        \addplot+[boxplot prepared={
            median=2,
            upper quartile=4,
            lower quartile=1,
            upper whisker=6,
            lower whisker=1
        }, fill, draw=black] coordinates {};
        \end{axis}
    \end{tikzpicture}
    \caption{\centering ``I believe this Showerthought has been written by a real person''}
    \label{fig:real_person}
\end{figure}

The box plots in Figures \ref{fig:general_score} -- \ref{fig:real_person} illustrate the evaluations provided by the survey participants regarding the Showerthoughts, grouped by statement and model.

\begin{figure}
    \centering
    \begin{tikzpicture}
        \begin{axis}[
            width=\linewidth,
            cycle list/Set2-8,
            boxplot/draw direction = y,
            x axis line style = {opacity=0},
            axis x line* = bottom,
            axis y line* = left,
            ymajorgrids,
            xtick = {1, 2, 3, 4},
            xticklabel style = {align=center, font=\scriptsize},
            xticklabels = {Genuine, GPT-Neo, ChatGPT, GPT-2},
            xtick style = {draw=none}, %
            ylabel = {Score},
            ytick = {1, 2, 3, 4, 5, 6}
        ]
        \addplot+[boxplot prepared={
            median=5,
            upper quartile=6,
            lower quartile=3,
            upper whisker=6,
            lower whisker=1
        }, fill, draw=black] coordinates {};
        \addplot+[boxplot prepared={
            median=4,
            upper quartile=6,
            lower quartile=3,
            upper whisker=6,
            lower whisker=1
        }, fill, draw=black] coordinates {};
        \addplot+[boxplot prepared={
            median=4,
            upper quartile=5,
            lower quartile=2,
            upper whisker=6,
            lower whisker=1
        }, fill, draw=black] coordinates {};
        \addplot+[boxplot prepared={
            median=3,
            upper quartile=5,
            lower quartile=1,
            upper whisker=6,
            lower whisker=1
        }, fill, draw=black] coordinates {};
        \end{axis}
    \end{tikzpicture}
    \caption{Logical Validity}
\end{figure}
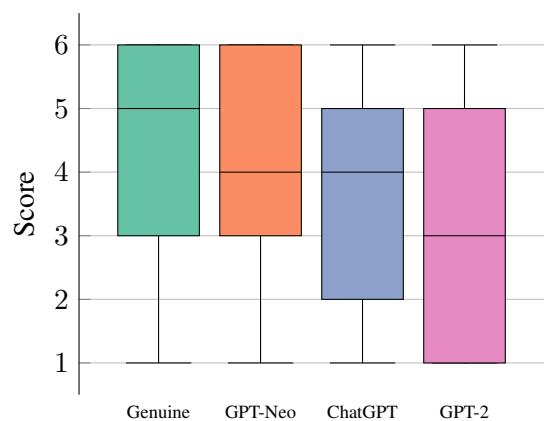

\begin{figure}[htbp]
    \centering
    \begin{tikzpicture}
        \begin{axis}[
            width=\linewidth,
            cycle list/Set2-8,
            boxplot/draw direction = y,
            x axis line style = {opacity=0},
            axis x line* = bottom,
            axis y line* = left,
            ymajorgrids,
            xtick = {1, 2, 3, 4},
            xticklabel style = {align=center, font=\scriptsize},
            xticklabels = {Genuine, GPT-Neo, ChatGPT, GPT-2},
            xtick style = {draw=none}, %
            ylabel = {Score},
            ytick = {1, 2, 3, 4, 5, 6}
        ]
        \addplot+[boxplot prepared={
            median=4,
            upper quartile=5,
            lower quartile=2,
            upper whisker=6,
            lower whisker=1
        }, fill, draw=black] coordinates {};
        \addplot+[boxplot prepared={
            median=3,
            upper quartile=5,
            lower quartile=2,
            upper whisker=6,
            lower whisker=1
        }, fill, draw=black] coordinates {};
        \addplot+[boxplot prepared={
            median=4,
            upper quartile=5,
            lower quartile=2,
            upper whisker=6,
            lower whisker=1
        }, fill, draw=black] coordinates {};
        \addplot+[boxplot prepared={
            median=2,
            upper quartile=4,
            lower quartile=1,
            upper whisker=6,
            lower whisker=1
        }, fill, draw=black] coordinates {};
        \end{axis}
    \end{tikzpicture}
    \caption{Creativity}
\end{figure}
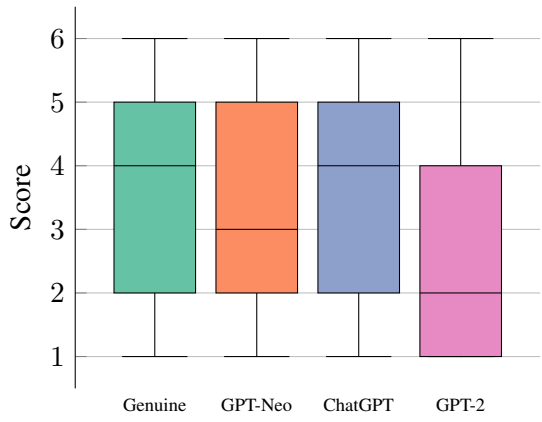

\begin{figure}
    \centering
    \begin{tikzpicture}
        \begin{axis}[
            width=\linewidth,
            cycle list/Set2-8,
            boxplot/draw direction = y,
            x axis line style = {opacity=0},
            axis x line* = bottom,
            axis y line* = left,
            ymajorgrids,
            xtick = {1, 2, 3, 4},
            xticklabel style = {align=center, font=\scriptsize},
            xticklabels = {Genuine, GPT-Neo, ChatGPT, GPT-2},
            xtick style = {draw=none}, %
            ylabel = {Score},
            ytick = {1, 2, 3, 4, 5, 6}
        ]
        \addplot+[boxplot prepared={
            median=3,
            upper quartile=5,
            lower quartile=2,
            upper whisker=6,
            lower whisker=1
        }, fill, draw=black] coordinates {};
        \addplot+[boxplot prepared={
            median=2,
            upper quartile=4,
            lower quartile=1,
            upper whisker=6,
            lower whisker=1
        }, fill, draw=black] coordinates {};
        \addplot+[boxplot prepared={
            median=3,
            upper quartile=4,
            lower quartile=1,
            upper whisker=6,
            lower whisker=1
        }, fill, draw=black] coordinates {};
        \addplot+[boxplot prepared={
            median=1,
            upper quartile=3,
            lower quartile=1,
            upper whisker=6,
            lower whisker=1
        }, fill, draw=black] coordinates {};
        \end{axis}
    \end{tikzpicture}
    \caption{Funniness}
\end{figure}
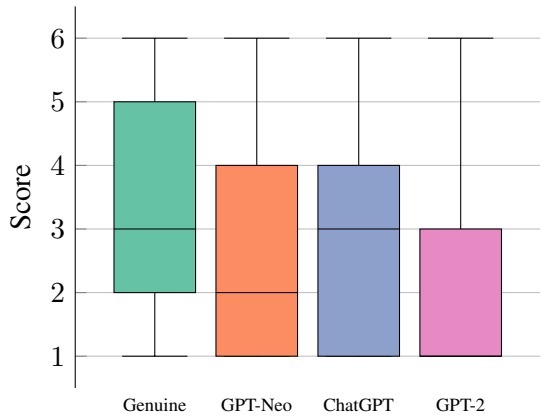

\begin{figure}[htbp]
  \centering
    \centering
    \begin{tikzpicture}
        \begin{axis}[
            width=\linewidth,
            cycle list/Set2-8,
            boxplot/draw direction = y,
            x axis line style = {opacity=0},
            axis x line* = bottom,
            axis y line* = left,
            ymajorgrids,
            xtick = {1, 2, 3, 4},
            xticklabel style = {align=center, font=\scriptsize},
            xticklabels = {Genuine, GPT-Neo, ChatGPT, GPT-2},
            xtick style = {draw=none}, %
            ylabel = {Score},
            ytick = {1, 2, 3, 4, 5, 6}
        ]
        \addplot+[boxplot prepared={
            median=4,
            upper quartile=5,
            lower quartile=2,
            upper whisker=6,
            lower whisker=1
        }, fill, draw=black] coordinates {};
        \addplot+[boxplot prepared={
            median=3,
            upper quartile=5,
            lower quartile=2,
            upper whisker=6,
            lower whisker=1
        }, fill, draw=black] coordinates {};
        \addplot+[boxplot prepared={
            median=3,
            upper quartile=4,
            lower quartile=2,
            upper whisker=6,
            lower whisker=1
        }, fill, draw=black] coordinates {};
        \addplot+[boxplot prepared={
            median=2,
            upper quartile=3,
            lower quartile=1,
            upper whisker=6,
            lower whisker=1
        }, fill, draw=black] coordinates {};
        \end{axis}
    \end{tikzpicture}
    \caption{Cleverness}
\end{figure}
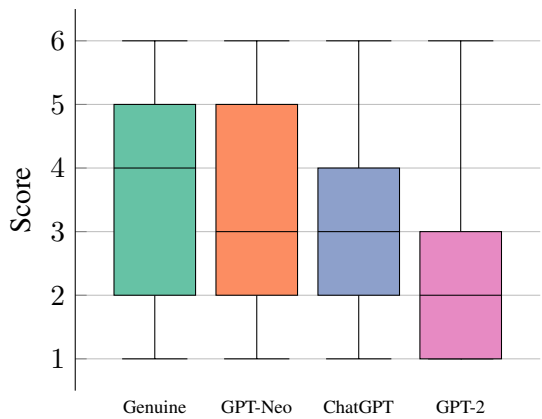

\subsection{RoBERTa Interpretability Results}

Figures \ref{figure:contributors_norm_training_gpt2} -- \ref{figure:contributors_norm_training_chatGpt_misclassified} depict which tokens had the highest influence towards the predicted class.

For instance, Figures~\ref{figure:contributors_norm_training_gpt2_generated}, \ref{figure:contributors_norm_training_gptNeo_generated}, and \ref{figure:contributors_norm_training_chatGpt_generated} depict the most significant contributors for predicting the generated class in the training data for each of the model-specific RoBERTa classifiers.

For an additional perspective, Figures~\ref{figure:contributors_norm_training_gpt2_missclassified}, \ref{figure:contributors_norm_training_gptNeo_missclassified}, and \ref{figure:contributors_norm_training_chatGpt_misclassified} show the most relevant contributors for misclassified Showerthoughts, i.e., the features that influenced the respective RoBERTa classifier to predict the wrong class.

\begin{figure*}[htbp]
    \centering
    \begin{subfigure}[b]{0.45\textwidth}
        \centering
        \begin{tikzpicture}
            \begin{axis} [
                width=\linewidth,
                xbar,
                ytick=data,
                symbolic y coords = {A, There, It, You, We, { }, {.}, The, {,}, If},
                xticklabel style={/pgf/number format/fixed, /pgf/number format/precision=5},
                scaled x ticks=false
            ]
            \addplot[fill=blue!30, draw=blue] coordinates {
                (0.043,If)
                (0.038,{,})
                (0.028,The)
                (0.022,{.})
                (0.015,{ })
                (0.011,We)
                (0.01,You)
                (0.01,It)
                (0.01,There)
                (0.01,A)
            };
            \end{axis}
        \end{tikzpicture}
        \caption{Predicted ``generated'' class}
        \label{figure:contributors_norm_training_gpt2_generated}
    \end{subfigure}
    \hfill
    \begin{subfigure}[b]{0.45\textwidth}
        \centering
        \begin{tikzpicture}
            \begin{axis} [
                width=\linewidth,
                xbar,
                ytick=data,
                symbolic y coords = {and, is, L, to, {,}, you, the, a, {.}, { }},
                xticklabel style={/pgf/number format/fixed, /pgf/number format/precision=5},
                scaled x ticks=false
            ]
            \addplot[fill=blue!30, draw=blue] coordinates {
                (0.019,{ })
                (0.015,{.})
                (0.006,a)
                (0.0055,the)
                (0.005,you)
                (0.0048,{,})
                (0.0046,to)
                (0.0042,L)
                (0.0038,is)
                (0.0034,and)
            };
            \end{axis}
        \end{tikzpicture}
        \caption{Predicted ``genuine'' class}
        \label{figure:contributors_norm_training_gpt2_genuine}
    \end{subfigure}
    \caption{Tokens with highest attribution scores towards the predicted class (GPT-2)}
    \label{figure:contributors_norm_training_gpt2}
\end{figure*}

\begin{figure*}[htbp]
    \centering
    \begin{subfigure}[b]{0.45\textwidth}
        \centering
        \begin{tikzpicture}
            \begin{axis} [
                width=\linewidth,
                xbar,
                ytick=data,
                symbolic y coords = {People, Every, It, A, When, There, We, You, If, The},
                xticklabel style={/pgf/number format/fixed, /pgf/number format/precision=5},
                scaled x ticks=false
            ]
            \addplot[fill=blue!30, draw=blue] coordinates {
                (0.049,The)
                (0.048,If)
                (0.022,You)
                (0.015,We)
                (0.013,There)
                (0.011,When)
                (0.009,A)
                (0.008,It)
                (0.0065,Every)
                (0.006,People)
            };
            \end{axis}
        \end{tikzpicture}
        \caption{Predicted ``generated'' class}
        \label{figure:contributors_norm_training_gptNeo_generated}
    \end{subfigure}
    \hfill
    \begin{subfigure}[b]{0.45\textwidth}
        \centering
        \begin{tikzpicture}
            \begin{axis} [
                width=\linewidth,
                xbar,
                ytick=data,
                symbolic y coords = {for, of, a, { }, people, is, the, in, {,}, {.}},
                xticklabel style={/pgf/number format/fixed, /pgf/number format/precision=5},
                scaled x ticks=false
            ]
            \addplot[fill=blue!30, draw=blue] coordinates {
                (0.0125,{.})
                (0.008,{,})
                (0.005,in)
                (0.0048,the)
                (0.004,is)
                (0.0038,people)
                (0.0036,{ })
                (0.0036,a)
                (0.0036,of)
                (0.003,for)
            };
            \end{axis}
        \end{tikzpicture}
        \caption{Predicted ``genuine'' class}
        \label{figure:contributors_norm_training_gptNeo_genuine}
    \end{subfigure}
    \caption{Tokens with highest attribution scores towards the predicted class (GPT-Neo)}
    \label{figure:contributors_norm_training_gptNeo}
\end{figure*}

\begin{figure*}[htbp]
    \centering
    \begin{subfigure}[b]{0.45\textwidth}
        \centering
        \begin{tikzpicture}
            \begin{axis} [
                width=\linewidth,
                xbar,
                ytick=data,
                symbolic y coords = {When, It, The, Why, {.}, is, {,}, { }, {?}, If},
                xticklabel style={/pgf/number format/fixed, /pgf/number format/precision=5},
                scaled x ticks=false
            ]
            \addplot[fill=blue!30, draw=blue] coordinates {
                (0.045,If)
                (0.0275,{?})
                (0.015,{ })
                (0.013,{,})
                (0.011,is)
                (0.01,{.})
                (0.009,Why)
                (0.008,The)
                (0.007,It)
                (0.006,When)
            };
            \end{axis}
        \end{tikzpicture}
        \caption{Predicted ``generated'' class}
        \label{figure:contributors_norm_training_chatGpt_generated}
    \end{subfigure}
    \hfill
    \begin{subfigure}[b]{0.45\textwidth}
        \centering
        \begin{tikzpicture}
            \begin{axis} [
                width=\linewidth,
                xbar,
                ytick=data,
                symbolic y coords = {don, a, L, the, you, is, The, are, { }, {.}},
                xticklabel style={/pgf/number format/fixed, /pgf/number format/precision=5},
                scaled x ticks=false
            ]
            \addplot[fill=blue!30, draw=blue] coordinates {
                (0.021,{.})
                (0.015,{ })
                (0.005,are)
                (0.0045,The)
                (0.004,is)
                (0.0035,you)
                (0.0035,the)
                (0.003,L)
                (0.0025,a)
                (0.002,don)
            };
            \end{axis}
        \end{tikzpicture}
        \caption{Predicted ``genuine'' class}
        \label{figure:contributors_norm_training_chatGpt_genuine}
    \end{subfigure}
    \caption{Tokens with highest attribution scores towards the predicted class (ChatGPT)}
    \label{figure:contributors_norm_training_chatGpt}
\end{figure*}

\begin{figure*}[htbp]
    \centering
    \begin{subfigure}[b]{0.45\textwidth}
        \centering
        \begin{tikzpicture}
            \begin{axis} [
                width=\linewidth,
                xbar,
                ytick=data,
                symbolic y coords = {more, is, of, as, We, A, {.}, There, The, You},
                xticklabel style={/pgf/number format/fixed, /pgf/number format/precision=5},
                scaled x ticks=false
            ]
            \addplot[fill=red!30, draw=red] coordinates {
                (0.021,You)
                (0.017,The)
                (0.014,There)
                (0.011,{.})
                (0.01,A)
                (0.008,We)
                (0.0075,as)
                (0.0075,of)
                (0.006,is)
                (0.0055,more)
            };
            \end{axis}
        \end{tikzpicture}
        \caption{Predicted ``generated'' class}
        \label{figure:contributors_norm_training_gpt2_missclassified_generated}
    \end{subfigure}
    \hfill
    \begin{subfigure}[b]{0.45\textwidth}
        \centering
        \begin{tikzpicture}
            \begin{axis} [
                width=\linewidth,
                xbar,
                ytick=data,
                symbolic y coords = {to, can, your, have, the, L, you, {.}, { }, a},
                xticklabel style={/pgf/number format/fixed, /pgf/number format/precision=5},
                scaled x ticks=false
            ]
            \addplot[fill=red!30, draw=red] coordinates {
                (0.011,a)
                (0.0105,{ })
                (0.009,{.})
                (0.0065,you)
                (0.0055,L)
                (0.004,the)
                (0.0035,have)
                (0.003,your)
                (0.0025,can)
                (0.002,to)
            };
            \end{axis}
        \end{tikzpicture}
        \caption{Predicted ``genuine'' class}
        \label{figure:contributors_norm_training_gpt2_missclassified_genuine}
    \end{subfigure}
    \caption{Tokens with highest attribution scores towards the predicted class when misclassified (GPT-2)}
    \label{figure:contributors_norm_training_gpt2_missclassified}
\end{figure*}

\begin{figure*}[htbp]
    \centering
    \begin{subfigure}[b]{0.45\textwidth}
        \centering
        \begin{tikzpicture}
            \begin{axis} [
                width=\linewidth,
                xbar,
                ytick=data,
                symbolic y coords = {A, Every, {,}, It, People, There, When, The, We, You},
                xticklabel style={/pgf/number format/fixed, /pgf/number format/precision=5},
                scaled x ticks=false
            ]
            \addplot[fill=red!30, draw=red] coordinates {
                (0.025,You)
                (0.017,We)
                (0.015,The)
                (0.014,When)
                (0.009,There)
                (0.008,People)
                (0.007,It)
                (0.006,{,})
                (0.0055,Every)
                (0.005,A)
            };
            \end{axis}
        \end{tikzpicture}
        \caption{Predicted ``generated'' class}
        \label{figure:contributors_norm_training_gptNeo_missclassified_generated}
    \end{subfigure}
    \hfill
    \begin{subfigure}[b]{0.45\textwidth}
        \centering
        \begin{tikzpicture}
            \begin{axis} [
                width=\linewidth,
                xbar,
                ytick=data,
                symbolic y coords = {have, The, how, can, on, {.}, who, people, {,}, { }},
                xticklabel style={/pgf/number format/fixed, /pgf/number format/precision=5},
                scaled x ticks=false
            ]
            \addplot[fill=red!30, draw=red] coordinates {
                (0.0068,{ })
                (0.0057,{,})
                (0.0045,people)
                (0.0044,who)
                (0.0042,{.})
                (0.0039,on)
                (0.0038,can)
                (0.0032,how)
                (0.003,The)
                (0.003,have)
            };
            \end{axis}
        \end{tikzpicture}
        \caption{Predicted ``genuine'' class}
        \label{figure:contributors_norm_training_gptNeo_missclassified_genuine}
    \end{subfigure}
    \caption{Tokens with highest attribution scores towards the predicted class when misclassified (GPT-Neo)}
    \label{figure:contributors_norm_training_gptNeo_missclassified}
\end{figure*}

\begin{figure*}[htbp]
    \centering
    \begin{subfigure}[b]{0.45\textwidth}
        \centering
        \begin{tikzpicture}
            \begin{axis} [
                width=\linewidth,
                xbar,
                ytick=data,
                symbolic y coords = {ant, on, were, When, someone, Old, a, de, L, ac},
                xticklabel style={/pgf/number format/fixed, /pgf/number format/precision=5},
                scaled x ticks=false
            ]
            \addplot[fill=red!30, draw=red] coordinates {
                (0.048,ac)
                (0.04,L)
                (0.038,de)
                (0.038,a)
                (0.035,Old)
                (0.03,someone)
                (0.027,When)
                (0.025,were)
                (0.023,on)
                (0.022,ant)
            };
            \end{axis}
        \end{tikzpicture}
        \caption{Predicted ``generated'' class}
        \label{figure:contributors_norm_training_chatGpt_missclassified_generated}
    \end{subfigure}
    \hfill
    \begin{subfigure}[b]{0.45\textwidth}
        \centering
        \begin{tikzpicture}
            \begin{axis} [
                width=\linewidth,
                xbar,
                ytick=data,
                symbolic y coords = {We, aG, world, {,}, soup, we, is, you, { }, {.}},
                xticklabel style={/pgf/number format/fixed, /pgf/number format/precision=5},
                scaled x ticks=false
            ]
            \addplot[fill=red!30, draw=red] coordinates {
                (0.048,{.})
                (0.043,{ })
                (0.009,you)
                (0.006,is)
                (0.004,we)
                (0.003,soup)
                (0.0028,{,})
                (0.0027,world)
                (0.0025,aG)
                (0.0023,We)
            };
            \end{axis}
        \end{tikzpicture}
        \caption{Predicted ``genuine'' class}
        \label{figure:contributors_norm_training_chatGpt_missclassified_genuine}
    \end{subfigure}
    \caption{Tokens with highest attribution scores towards the predicted class when misclassified (ChatGPT)}
    \label{figure:contributors_norm_training_chatGpt_misclassified}
\end{figure*}

\end{document}